\documentclass[runningheads]{llncs}

\usepackage[mobile]{eccv}
\usepackage[dvipsnames]{xcolor}
\usepackage{graphicx, calc}
\usepackage{amssymb}
\usepackage{booktabs}
\usepackage[linesnumbered,ruled,vlined]{algorithm2e}
\usepackage{multirow}
\usepackage{makecell}
\usepackage[export]{adjustbox}
\usepackage{xcolor}
\usepackage{enumitem}
\usepackage{tikz}
\usepackage{colortbl}
\usepackage{color,soul}
\usepackage{cancel}

\usepackage{enumitem}

\usepackage{subcaption}
\usepackage{comment}
\usepackage{tabularx}
\usepackage{pgffor} %
\usepackage{etoolbox}
\usepackage{colortbl}

\usepackage{xifthen}
\usepackage{multirow}
\usepackage{rotating}
\usepackage{booktabs}
\usepackage{array}
\usepackage[export]{adjustbox}
\usepackage{arydshln}
\usepackage{tcolorbox}
\usepackage{nicefrac}

\newcommand{\ASourceOut}{\textcolor{source}{$\mathbb{A}_\text{Source}^\text{\textbf{\textit{Out}}}$}}
\newcommand{\AWebOut}{\textcolor{web}{$\mathbb{A}_\text{Web}^\text{\textbf{\textit{Out}}}$}}
\newcommand{\ASyntheticOut}{\textcolor{synthetic}{$\mathbb{A}_\text{Synthetic}^\text{\textbf{\textit{Out}}}$}}
\newcommand{\ASourceIn}{\textcolor{source}{$\mathbb{A}_\text{Source}^\text{\textbf{\textit{In}}}$}}
\newcommand{\AWebIn}{\textcolor{web}{$\mathbb{A}_\text{Web}^\text{\textbf{\textit{In}}}$}}
\newcommand{\ASyntheticIn}{\textcolor{synthetic}{$\mathbb{A}_\text{Synthetic}^\text{\textbf{\textit{In}}}$}}

\definecolor{darkgreen}{RGB}{0,100,0}  %
\definecolor{darkred}{RGB}{139,0,0}    %

\newcolumntype{Y}[1]{>{\centering\arraybackslash\hspace{0pt}}p{#1\textwidth}}
\newcolumntype{H}{>{\setbox0=\hbox\bgroup}c<{\egroup}@{}}
\newcolumntype{C}[1]{>{\centering\arraybackslash}p{#1}}
\newcolumntype{E}{>{\centering\arraybackslash}X}

\newcommand{\gainloss}[1]{%
	\pgfmathsetmacro\result{sign(#1)}%
	\ifdim\result pt=1pt %
	{\scriptsize \textcolor{green}{(+#1)}}%
	\else
	\ifdim\result pt=-1pt %
	{\scriptsize \textcolor{red}{(#1)}}%
	\else
	\scriptsize{(+#1)} %
	\fi
	\fi
}

\makeatletter
\def\adl@drawiv#1#2#3{%
        \hskip.5\tabcolsep
        \xleaders#3{#2.5\@tempdimb #1{1}#2.5\@tempdimb}%
                #2\z@ plus1fil minus1fil\relax
        \hskip.5\tabcolsep}
\newcommand{\cdashlinelr}[1]{%
  \noalign{\vskip\aboverulesep
           \global\let\@dashdrawstore\adl@draw
           \global\let\adl@draw\adl@drawiv}
  \cdashline{#1}
  \noalign{\global\let\adl@draw\@dashdrawstore
           \vskip\belowrulesep}}
\makeatother

\newtcolorbox[auto counter]{conclusions}[1][]{
  title={\bfseries Insight (\thetcbcounter)},
  coltitle=white,
  #1
}
\usepackage{multicol}

\usepackage{amsmath}

\newif\ifcvpr

\newif\ifeccv

\eccvtrue

\makeatletter
\NewDocumentCommand{\MakeTitleInner}{ +m +m +m }{
    \newpage%
    \null%
    \vskip 2em%
    \begin{center}%
        \let \footnote \thanks
        {\LARGE #1 \par}%
        \vskip 1.5em%
        {%
            \large
            \lineskip .5em%
            \begin{tabular}[t]{c}%
                #2%
            \end{tabular}\par%
        }%
        \vskip 1em%
        {\large #3}%
    \end{center}%
    \par
    \vskip 1.5em%
}
\NewDocumentCommand{\MakeTitle}{ +m +m +m }{%
    \begingroup
        \renewcommand\thefootnote{\@fnsymbol\c@footnote}%
        \def\@makefnmark{\rlap{\@textsuperscript{\normalfont\@thefnmark}}}%
        \long\def\@makefntext##1{\parindent 1em\noindent
            \hb@xt@1.8em{%
                \hss\@textsuperscript{\normalfont\@thefnmark}%
            }##1%
        }%
        \if@twocolumn
            \ifnum \col@number=\@ne
                \MakeTitleInner{#1}{#2}{#3}
            \else
                \twocolumn[\MakeTitleInner{#1}{#2}{#3}]%
            \fi
        \else
            \newpage
            \global\@topnum\z@   %
            \MakeTitleInner{#1}{#2}{#3}
        \fi
        \thispagestyle{plain}\@thanks
    \endgroup
    \setcounter{footnote}{0}%
}
\makeatother

\usepackage{eccvabbrv}
\usepackage{graphicx}
\usepackage{booktabs}

\usepackage[accsupp]{axessibility}  %

\usepackage[breaklinks,colorlinks,citecolor=eccvblue]{hyperref}

\usepackage{orcidlink}

\definecolor{source}{HTML}{e07000}
\definecolor{synthetic}{HTML}{076db5}
\definecolor{web}{HTML}{205f5b}

\begin{document}

\title{On Pretraining Data Diversity\\for Self-Supervised Learning}
\titlerunning{On Pretraining Data Diversity for SSL}

\author{Hasan Abed Al Kader Hammoud\inst{1,2}$^{\star\diamond}$ \and
Tuhin Das\inst{2}$^{\star}$ \and
    Fabio Pizzati\inst{2}\thanks{Equal contribution $^{\dagger}$ Equal supervision $^\diamond$ Work done during a research visit at Oxford} \\  Philip H.S. Torr\inst{2} \and Adel Bibi\inst{2}$^{\dagger}$ \and Bernard Ghanem\inst{1}$^{\dagger}$}

\authorrunning{Hammoud et al.}

\institute{KAUST \and University of Oxford}

\maketitle

\begin{abstract}

We explore the impact of training with more diverse datasets, characterized by the number of unique samples, on the performance of self-supervised learning (SSL) under a fixed computational budget. Our findings demonstrate that increasing pretraining data diversity enhances SSL performance, albeit only when the distribution distance to the downstream data is minimal. Notably, even with an exceptionally large pretraining data diversity achieved through methods like web crawling or diffusion-generated data, among other ways, the distribution shift remains a challenge. Our experiments are comprehensive with seven SSL methods using large-scale datasets such as ImageNet and YFCC100M amounting to over 200 GPU days. The code and trained models are available at \small{\url{https://github.com/hammoudhasan/DiversitySSL}.}\looseness=-1

\end{abstract}
\begin{figure}[t]
	\centering
    \includegraphics[width=0.85\linewidth]{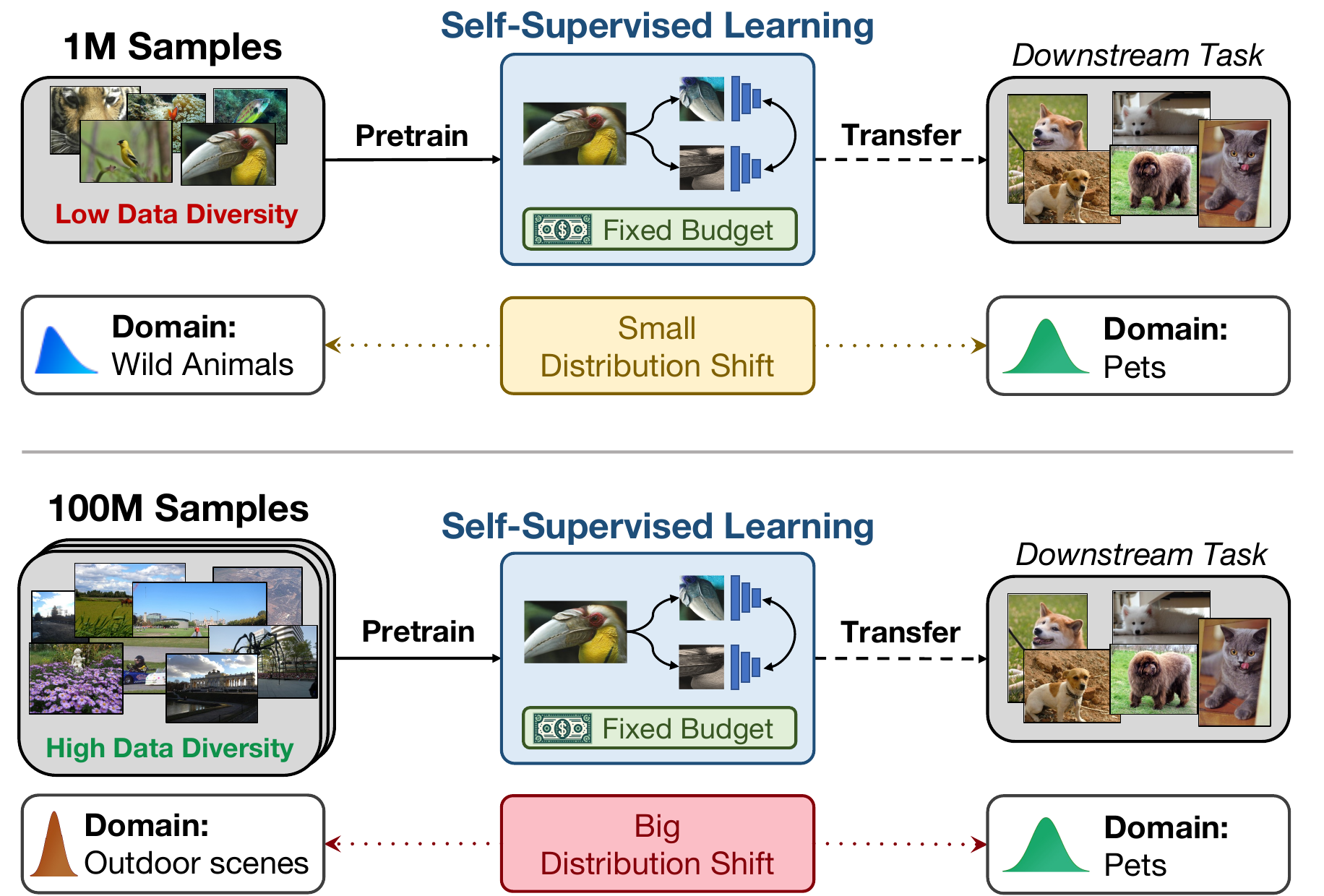}
    \caption{\textbf{Impact of Pretraining Diversity:} Self-supervised learning (SSL) can be used to pretrain vision models on small datasets closely aligned to the downstream task, \eg, pets classification, hence with a small distribution shift (top, wild animals pretraining). Conversely, we could pretrain on an extensively varied dataset, with wide distribution differences (outdoor scenes, bottom). We study the role of pretraining diversity in SSL under a fixed budget, and highlight its effects in relationship to the distribution shift.\looseness=-1}
	\label{fig:teaser} 
\end{figure}

\section{Introduction}
\label{sec:intro}
Self-supervised learning (SSL) has recently emerged as a new paradigm to pretrain large vision models at scale~\cite{goyal2021self,oquab2023dinov2, goyal2022vision}. Leveraging the ability to learn from unlabelled data, pretraining on millions---or even billions~\cite{sun2017revisiting, goyal2022vision}---of images turned from an unachievable goal to a common practice. This exposure to extremely \textit{diverse} datasets, \ie, 
composed of a remarkable number of unique samples, granted impressive performance and unprecedented generalization capabilities to a growing number of vision models~\cite{oquab2023dinov2}.
Large-scale datasets, in conjunction with substantial computational resources, have been the key driving forces for the success of SSL-based approaches. For instance, SEER~\cite{goyal2021self}, pretrained for approximately 11 GPU years on a billion images, exemplifies the massive computation and data resources employed for these models.
It has become the implicit norm that increasing computation and data is beneficial, without any detailed analysis of how they separately impact SSL effectiveness. In particular, it is not clear to what extent large datasets are responsible for the impressive generalization capabilities of SSL models. Indeed, consider the example in Figure~\ref{fig:teaser}. Assuming a fixed monetary budget, in the form of computational expenses, for pretraining a vision model under the SSL paradigm, does iterating over a large set of images work best given a downstream task, or it is better to train repeatedly on a smaller set of samples visually close to the ones of the downstream? In this context, we face the ever-lasting problem of
distribution shift on vision models~\cite{torralba2011unbiased}.
Without a proper understanding of the impact of data, there could be a wrong allocation of efforts in increasing data or computation, leading to inefficiencies in the deployment process of SSL models.

In this paper, we study the role of the diversity of the SSL pretraining data, under a fixed computational budget, when the pretraining data matches the distribution of the downstream data, as well as when they differ. Our experiments span various SSL methods~\cite{simclr,mocov3,vicreg,dino,byol,barlow,swav} and multiple datasets~\cite{cifar100, tinyimagenet, imagenet100, imagenet, thomeeYFCC100MNewData2016, stanfordcars, flowers102, pets, places365, food101}, and are tested under several computational budgets amounting to a total of 200 GPU days. We summarize our contributions below:

\begin{enumerate}[topsep=0pt,itemsep=0pt,parsep=0pt,partopsep=0pt]
    \item We show that SSL pretraining strategies are currently data-inefficient in compensating for distribution shifts. Under normalized computational costs, we verify that pretraining on large datasets with high diversity cannot outperform the pretraining on datasets with limited diversity, but with a lower distribution shift with respect to the downstream tasks. %
    \item %
    We conclude that there is a wide margin for improvement in the performance of current SSL methods on out-of-distribution classes, and we propose insights for a fair evaluation. This shows the need for future research on new SSL techniques leveraging data diversity better, to improve generalization capabilities beyond the training data distribution.
    \item We propose a novel strategy for a computationally-normalized evaluation of SSL and a carefully designed set of experiments focusing on pretraining data diversity, enabling us to draw our novel conclusions.
\end{enumerate}

\noindent Ultimately, our work provides a comprehensive analysis, from experiments worth 200 GPU days on NVIDIA A100, on the interaction between data diversity and computational resources, and their impact on the performance of SSL models.%

\section{Related Work}
\label{sec:relatedwork}

\noindent\textbf{Self-Supervised Learning\ \ }
Early work in self-supervised learning used simple pretext tasks, such as relative patch prediction~\cite{doersch2015unsupervised, doersch2017multi}, image colorization~\cite{zhangColorfulImageColorization2016}, image rotation prediction~\cite{gidarisUnsupervisedRepresentationLearning2018}, or solving jigsaw puzzles~\cite{norooziUnsupervisedLearningVisual2016} to train feature extractors in absence of annotations.
Representations learned with those methods are limitedly effective, hence more recent literature has moved towards more sophisticated approaches, such as using image augmentations to generate correlated views of a training sample, and learning to extract augmentation-invariant representations for these correlated pairs. Among these multi-view methods, many exploit contrastive losses~\cite{swav, simclr, oord2018representation, henaffDataEfficientImageRecognition2019, chenBigSelfSupervisedModels2020, chen2020improved, heMomentumContrastUnsupervised2020, liPrototypicalContrastiveLearning2021, bachman2019learning, hjelm2018learning, misra2020self}, enforcing similarity between views of the same image (positives) and dissimilarity from other images (negatives).
Due to the need of many negative samples, contrastive methods often require large batch sizes to work effectively~\cite{simclr, mocov3}.
Cluster-based methods such as SwAV~\cite{swav}, DINO~\cite{dino} and DeepCluster v2~\cite{caronDeepCluster} learn generalizable representations by grouping samples into cluster prototypes. Others exploit prediction of features with siamese networks~\cite{chen2021exploring}, learn features in a teacher-student fashion~\cite{byol}, or use redundancy reduction techniques~\cite{barlow, vicreg}. Finally, masked image modeling~\cite{bao2022beit, he2022masked, xie2022simmim} emerged as a scalable alternative for Vision Transformers~\cite{dosovitskiyImageWorth16x162021}, learning representations by predicting masked image patches as SSL task.\\

\noindent\textbf{Pretraining at Scale\ \ } SSL pretraining is most effective at scale, benefiting from large datasets and great computational resources. Initial attempts at large-scale pretraining were made by combining contrastive learning and clustering~\cite{tianDivideContrastSelfsupervised2021, caronUnsupervisedPreTrainingImage2019}. SEER~\cite{goyal2021self} was trained on 1 billion internet images with SwAV~\cite{swav} on a RegNet~\cite{radosavovic2020designing} backbone.
The importance of model scaling to leverage large datasets is also shown in ~\cite{xieDataScalingMasked2023, goyal2019scaling, goyal2022vision}, as well as the need for longer training times~\cite{xieDataScalingMasked2023}. Additional strategies are necessary to achieve best performance~\cite{zhai2022scaling, dehghani2023scaling, fini2023improved}. These works focus on reaching the best representations, without considerations about training costs, thus encouraging unfair comparisons across trainings. Preliminary papers~\cite{coleWhenDoesContrastive2022,el-noubyAreLargescaleDatasets2021} found that increasing data may lead to a modest boost in downstream performance when the downstream distribution changes. While they give insights on the pretraining diversity, its relation with the distribution shift under normalized computation is still not investigated.\\

\noindent\textbf{Distribution Shift in SSL\ \ } 
Some studies have investigated how the pretraining data affects downstream performance on other domains. In general, the distribution shift increases when factors such as visual content and data acquisition procedure differ in two data ensembles~\cite{torralba2011unbiased}.
For various pretraining datasets, preliminary works~\cite{kotar2021contrasting, coleWhenDoesContrastive2022, goyal2019scaling, li2023internet} observed that the best performing representations were learned on datasets that were similar to the downstream test task.
Additionally, combining datasets before pretraining or combining self-supervised features learned from various datasets did not lead to significant improvements in~\cite{coleWhenDoesContrastive2022}.
In~\cite{zhao2021what,shi2022robust}, they showcase higher generalization capabilities of SSL models compared to their supervised counterparts, for several downstream tasks in the presence of a distribution shift. In~\cite{van2021revisiting}, they pretrained on several datasets, observing different generalization depending on the appearance of the pretraining dataset. Furthermore, initial considerations on the impact of external data on downstream tasks under distribution shift have been proposed in~\cite{van2021benchmarking}. Some compensate the distribution shift with the scale of training datasets~\cite{hammoud2024synthclip}. However, there is still a lack of a fair, computation-normalized evaluation that allows to study the effects of the distribution shift in a controlled environment.

\section{Preliminaries}
\label{sec:prelims}

\noindent\textbf{Pretraining\ \ } 
We first outline the general pretraining procedure common to state-of-the-art self-supervised learning %
methods. Specific differences among these methods are detailed in the supplementary material. The overall pretraining pipeline, common across many SSL approaches~\cite{simclr,byol,barlow,dino,swav,vicreg}, goes as follows: (1) \textbf{Data Sampling:} from a large-scale, unlabeled \textit{upstream} pretraining dataset, denoted as $\mathbb{D}_{\text{SSL}}$, an image $\mathbf{x}$ is randomly sampled; (2) \textbf{View Generation:} two correlated views, $\Tilde{\mathbf{x}}_A$ and $\Tilde{\mathbf{x}}_B$, are generated from $\mathbf{x}$ using two image augmentation functions sampled at random. These random augmentations include random resized cropping, horizontal flipping, blurring, and color adjustments~\cite{cookbook}, among others;
(3) \textbf{Feature Extraction and Projection:} the correlated views undergo feature extraction through a network, $f_{\theta_f}$, parameterized by $\theta_f$, such as a ResNet~\cite{heDeepResidualLearning2016} or a ViT~\cite{dosovitskiyImageWorth16x162021}, leading to representations $\mathbf{h}_A=f_{\theta_f}(\tilde{\mathbf{x}}_A)$ and $\mathbf{h}_B=f_{\theta_f}(\tilde{\mathbf{x}}_B)$. A linear projection head, $g_{\theta_g}$, parameterized by $\theta_g$, then maps these representations to a latent space, resulting in $\mathbf{z}_A = g_{\theta_g}(\mathbf{h}_A)$ and $\mathbf{z}_B = g_{\theta_g}(\mathbf{h}_B)$; (4) \textbf{Joint Optimization:} The feature extractor $f_{\theta_f}$ and the projection head $g_{\theta_g}$ are jointly optimized according to the following objective:

\begin{equation}
\theta_f^*, \theta_g^* = \underset{\theta_f, \theta_g}{\arg\min}~\mathbb{E}_{\mathbf{x} \sim \mathbb{D}_{\text{SSL}}}~\mathcal{L}_\text{SSL}(\mathbf{z}_A, \mathbf{z}_B),
\end{equation}

\noindent where $\mathcal{L}_\text{SSL}$ is a loss function specific to the SSL pretraining method. After pretraining, the feature extractor $f_{\theta_f^*}$ can be deployed for various \textit{downstream} tasks such as image classification, object detection, or segmentation. This is typically achieved by training a task-specific head. Alternatively, the feature extractor $f_{\theta_f^*}$ can be fine-tuned or used with a $k$-nearest neighbors (kNN) classifier.\\

\noindent\textbf{Linear Probing\ \ } There are several ways to evaluate the performance of a self-supervised learning method such as linear probing~\cite{byol, mocov3, simclr}, kNN~\cite{wu2018unsupervised, zhuang2019local, dino, swav}, and few-shot evaluation~\cite{goyal2021self, ericsson2021well}. %
Consistently with the general protocol established in the literature~\cite{simclr, heMomentumContrastUnsupervised2020, zhangColorfulImageColorization2016, coleWhenDoesContrastive2022}, we use linear probing to measure the quality of the features extracted for classification tasks. The procedure is:
(1) a labeled \textit{downstream} dataset, $\mathbb{D}_\text{task}$, consisting of image-class pairs $(\mathbf{x}, \mathbf{y})\sim\mathbb{D}_\text{task}$, is selected for evaluation. (2) For each image $\mathbf{x}$, its representation 
is extracted using the pretrained feature extractor $f_{\theta_f^*}$, after which the linear classification head $t_{\theta_t}$, parameterized by $\theta_t$, is then applied to obtain $t_{\theta_t}(f_{\theta_f^*}(\mathbf{x}))$.
(3) The linear head $t_{\theta_t}$ is optimized as follows:
        \begin{equation}
        \theta^{*}_t={\underset{\theta_t}{\arg\min}~\mathbb{E}_{(\mathbf{x},\mathbf{y}) \sim \mathbb{D}_{\text{task}}}\left[\mathcal{L}_\text{task}(t_{\theta_t}(f_{\theta_f^*}(\mathbf{x})), \mathbf{y})\right]},
        \label{eq:optimization}
    \end{equation}
Note that only the parameters of the head $\theta_t$ are optimized, while the parameters $\theta^{*}_f$ of the feature extractor are frozen. The quality of features extracted by $f_{\theta_f^*}$ is directly inferred from the classification accuracy achieved on the test set of $\mathbb{D}_\text{task}$, which serves as the primary indicator of the quality of the extracted features.\looseness=-1

\section{Normalized Evaluation} \label{sec:our_formulation}
We stress how for a correct understanding of the impact of data diversity we need to analyze its effects isolating them from the impact of increased computation. To enable this, we introduce (1) a computational budget used to normalize computation across experiments, and (2) a quantification of the data diversity seen by models during pretraining.\\

\noindent\textbf{Computational Budget\ \ }
Current progress in SSL pretraining simultaneously scales computational budget and dataset size to achieve the best performance~\cite{goyal2021self}.
This makes it difficult to assess the reasons behind the success of many SSL algorithms.
Do these methods benefit mainly from the large amounts of computation, \ie, running SSL pretraining for large numbers of epochs, or do they benefit from data diversity in larger datasets containing a vast variety of visual features?
To perform a disentangled evaluation of these two factors, we first introduce $\mathcal{C}$ as a measure of computational budget, which quantifies the total number of images an SSL method is allowed to process during pretraining. This is calculated as $\mathcal{C} = N \cdot \mathcal{E}$, where $N$ is the number of unique samples in the pretraining dataset $\mathbb{D}_\text{SSL}$, hence the \textit{data diversity} of the dataset, and $\mathcal{E}$ is the number of pretraining epochs. 
Constraining multiple models pretrained with SSL to a fixed computational budget $\mathcal{C}$ allows for meaningful comparison, as it is guaranteed that all SSL methods will have processed the same number of pretraining images.

\begin{table*}[t]
\caption{\textbf{Impact of Data Diversity on CIFAR-100 and Tiny ImageNet SSL Pretraining Performance}: We study the effects of diversity on CIFAR-100~\protect{\subref{tab:cifar-100}} and Tiny ImageNet~\protect{\subref{tab:tiny-imagenet}} across seven different methods and three data diversity settings for a ResNet-18 pretraining, where for all, $\mathbb{D}_\text{task}=\mathbb{D}_\text{SSL}$. This comparison includes analyzing classification accuracy through linear probing on the training set and evaluation on the test set of the respective datasets. Although performance fluctuates among different methods, a consistent trend is observed: higher data diversity typically leads to the generation of higher quality representations.}\label{tab:small-scale}
	\centering
	\begin{subtable}{0.495\linewidth}
        \Large
\setlength{\aboverulesep}{0pt}
\setlength{\belowrulesep}{0pt}\setlength\tabcolsep{0.18em} %
	\resizebox{\linewidth}{!}{%
		\begin{tabular}{cc|cC{1cm}C{1cm}C{1cm}C{1cm}C{1cm}C{1cm}C{1cm}}
		\toprule
		$N$ & $\mathcal{D}$ & \multicolumn{8}{c}{\textbf{Accuracy$\uparrow$}} \\
         $\times 10^3$ & $\times 10^{-3}$ &&  \small{SimCLR} & \small{B.T.} & \small{BYOL} & \small{SwAV} & \small{VICR} & \small{\mbox{MoCo3}} & \small{DINO}\\
		\midrule
		5  & 0.1 && 38.0 & 44.3&41.1&14.1&44.1& 12.5 &15.7\\
		25 & 0.5 && 50.1&54.3&51.0&45.4&50.4&51.6&35.9 \\
		50 & 1.0 && \textbf{58.6}&\textbf{58.4}&\textbf{58.3}&\textbf{56.4}&\textbf{55.8}&\textbf{55.6}&\textbf{40.9}\\
		\bottomrule
        \\[-1.5\medskipamount]
		\multicolumn{10}{c}{$\mathcal{C}=50\times 10^6$}
	\end{tabular}
	}
	\caption{CIFAR-100}\label{tab:cifar-100}
\end{subtable}	
\hfill
\begin{subtable}{0.495\linewidth}
\Large
\setlength\tabcolsep{0.18em} %
\setlength{\aboverulesep}{0pt}
\setlength{\belowrulesep}{0pt}
\resizebox{\linewidth}{!}{%
	\begin{tabular}{cc|cC{1cm}C{1cm}C{1cm}C{1cm}C{1cm}C{1cm}C{1cm}}
		\toprule
		$N$ & $\mathcal{D}$ & \multicolumn{8}{c}{\textbf{Accuracy$\uparrow$}} \\
         $\times 10^3$ &  $\times 10 ^{-3}$ &&  \small{SimCLR} & \small{B.T.} & \small{BYOL} & \small{SwAV} & \small{VICR} & \small{\mbox{MoCo3}} & \small{DINO}\\
        \midrule
		 10 & 0.2 && 36.9 & 41.0 & 35.6 & 34.2 & 37.6 & 36.6 & 34.5 \\
		 50 & 1.0 && 48.8 & 52.0 & 48.1 & 43.6 & 48.8 & 46.5 & 44.4 \\
		 100 & 2.0 && \textbf{49.8} & \textbf{55.6} & \textbf{50.3} & \textbf{47.5} & \textbf{54.1} & \textbf{48.6} & \textbf{47.3} \\
		\bottomrule
        \\[-1.5\medskipamount]
        \multicolumn{10}{c}{$\mathcal{C}=50\times 10^6$}

	\end{tabular}
}
	\caption{Tiny ImageNet}\label{tab:tiny-imagenet}
\end{subtable}

\end{table*}
\noindent\textbf{Quantifying Pretraining Diversity\ \ } Even under normalized computation $\mathcal{C}$, various SSL approaches may be exposed to different data diversity as they are trained with different datasets. For instance, a model trained for $\mathcal{E} = 1000$ epochs on a dataset of size $N=1000$ will see less diversity than a model pretrained for $\mathcal{E}=1$ epoch on a dataset of size $N=10^{6}$, despite that they are both pretrained under normalized computation, processing the same number of images. Hence, to capture this notion of exposure to diversity while pretraining under normalized computation $\mathcal{C}$, we define a \textit{pretraining diversity} $\mathcal{D}$, which captures the number of unique samples encountered during training given a fixed $\mathcal{C}$ as
$\mathcal{D} = \nicefrac{N}{\mathcal{C}} = \nicefrac{1}{\mathcal{E}}$. %
A model pretrained with larger $\mathcal{D}$ indicates that the model is presented with a larger number of unique images during training with fixed $\mathcal{C}$, and hence is exposed to more pretraining data diversity. In the next section, we explore the effects of variations in $\mathcal{D}$ on SSL performance under a distribution shift, while keeping $\mathcal{C}$ constant.

\section{Fixed Budget SSL: In \& Out-of-Distribution}\label{sec:experiments}

\noindent\textbf{Training Configuration\ \ } We evaluate seven SSL methods: SimCLR~\cite{simclr}, MoCoV3~\cite{mocov3}, VICReg~\cite{vicreg}, DINO~\cite{dino}, BYOL~\cite{byol}, Barlow Twins~\cite{barlow}, and SwAV~\cite{swav}. We use different datasets for both pretraining $\mathbb{D}_{\text{SSL}}$ and linear probing $\mathbb{D}_{\text{task}}$ for different sections.
We use \texttt{solo-learn}~\cite{costaSololearnLibrarySelfsupervised2022} as a codebase. For each method, we use the default parameters provided when available, otherwise, we conduct a hyperparameter search to ensure proper convergence. \looseness=-1
\subsection{Performance on the Same Distribution}\label{sec:exp-part1}

\noindent We aim to answer a first question: \textit{does collecting more samples from the same distribution help SSL pretraining with a fixed budget?} We conduct experiments to capture how the pretraining diversity $\mathcal{D}$ %
influences SSL pretraining within a normalized $\mathcal{C}$, focusing on the simplest setting where the upstream and downstream datasets belong to the same distribution such that $\mathbb{D}_{\text{SSL}} = \mathbb{D}_{\text{task}} $. This serves as a fundamental ground for our further experiments.

\noindent\textbf{Setup\ \ }
For pretraining, we use CIFAR-100~\cite{cifar100} and Tiny ImageNet~\cite{tinyimagenet} as \(\mathbb{D}_\text{SSL}\), which contain $50\times 10^3$ and $100\times 10^3$ images, respectively. We set \(\mathcal{C}=50\times 10^{6}\), chosen such that it allows all methods to converge during pretraining. In Section~\ref{sec:ablations}, we study the effect of varying the budget \(\mathcal{C}\).
We pretrain on subsets of \(\mathbb{D}_\text{SSL}\) with different sizes $N$ (10\%, 50\%, and 100\% of $\mathbb{D}_{\text{SSL}}$), enabling us to observe the effects of pretraining diversity 
$\mathcal{D}$ on training outcomes where $\mathcal{E}$ is adjusted accordingly to match the budget \(\mathcal{C}\). For example, using 100\% of CIFAR-100 involves 1000 %
epochs of training, while 10\% and 50\% of the dataset lead to 10000 and 2000 epochs, respectively. These subsets of \(\mathbb{D}_\text{SSL}\) are created by shuffling the dataset and then selecting the first \(10\%, 50\%\), or \(100\%\) split.
All models in this section use a ResNet-18~\cite{heDeepResidualLearning2016} backbone pretrained from scratch. %
For evaluation, we use in-distribution linear probing, \ie, \(\mathbb{D}_\text{task} = \mathbb{D}_\text{SSL}\), where performance is measured by classification accuracy on the test set.\\

\noindent\textbf{Results\ \ } The results of linear probing are presented in Tables~\ref{tab:cifar-100} and~\ref{tab:tiny-imagenet} for CIFAR-100 and Tiny ImageNet, respectively.
While different SSL methods show varying levels of performance, we observe that, for all methods, an increase in the diversity $\mathcal{D}$ consistently leads to higher linear probing accuracy, suggesting that including more in-distribution samples in $\mathbb{D}_\text{SSL}$ helps under a normalized $\mathcal{C}$.
For example, SimCLR achieves an accuracy of 37.95 when 10\% of CIFAR-100 is provided, whereas this performance improve by around 12\% when only 50\% of the dataset is provided and by another 8\% after pretraining on 100\% of unique samples. This suggests that SSL methods substantially benefit from having a more diverse pretraining set in computationally budgeted in-distribution evaluation, a fundamental verification that allows us to proceed in our analysis. 

\begin{conclusions}[colback=white]
When the distributions of the upstream and downstream tasks are the same, \ie, \(\mathbb{D}_\text{SSL}\) = \(\mathbb{D}_\text{task}\), in a normalized computation setup, increasing pretraining diversity $\mathcal{D}$ proves to be an effective strategy for enhancing SSL pretraining performance.
\end{conclusions}

\subsection{Increasing Data Diversity}\label{sec:exp-part2}

As observed in Section~\ref{sec:exp-part1}, if $\mathbb{D}_{\text{SSL}} = \mathbb{D}_{\text{task}}$ having a more diverse pretraining set benefits SSL methods, under a normalized computation assumption. %
However, to increase diversity, sampling from the same distribution $\mathbb{D}_{\text{task}}$ to extend $\mathbb{D}_{\text{SSL}}$ is not always attainable. In fact, pretraining data is often sampled from a distribution different than $\mathbb{D}_{\text{task}}$. The added samples will then introduce a distribution shift between $\mathbb{D}_\text{SSL}$ and $\mathbb{D}_\text{task}$. Then, we raise the following question: \textit{is increasing pretraining diversity still beneficial when there is a distribution shift between $\mathbb{D}_\text{SSL}$ and $\mathbb{D}_\text{task}$, \ie, $\mathbb{D}_{\text{SSL}} \neq \mathbb{D}_{\text{task}}$}?  We explore strategies for acquiring new data for increasing $\mathcal{D}$, namely, including existing samples, crawled internet data, and data synthetically generated. To evaluate the effects of the distribution shift in a controlled scenario, we analyze distributions closer to $\mathbb{D}_\text{task}$ by using a class prior (in-distribution) and without a class prior (out-of-distribution).

\begin{figure}[t]
    \centering
    \includegraphics[width=0.9\linewidth]{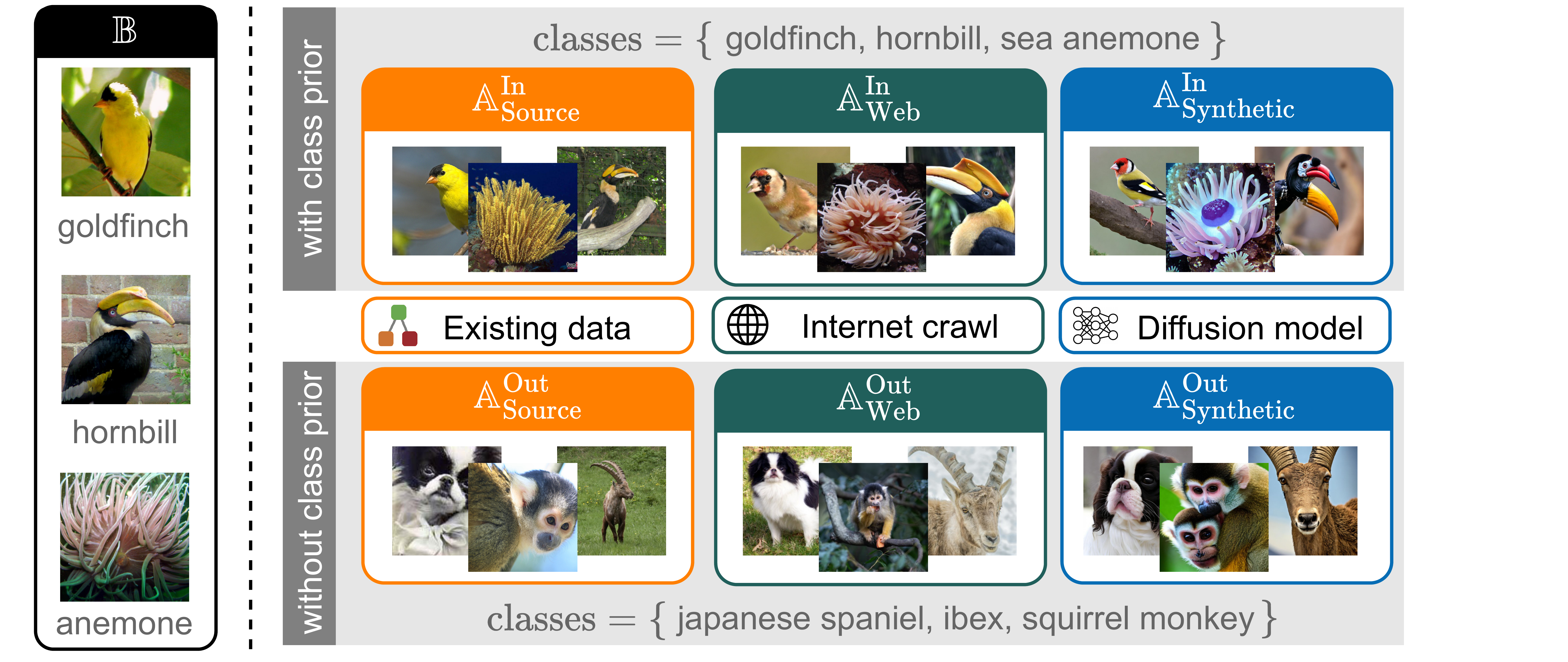}
    \caption{\textbf{Data Collection Strategies: } We analyze strategies for collecting additional data ($\mathbb{A}$), \ie, collecting more \textcolor{source}{source} data, crawling the \textcolor{web}{web} or using \textcolor{synthetic}{synthetic} images. Using a class prior (top row) simulates \textbf{\textit{In-}}distribution trainings. We also collect images without class prior (bottom row) to analyze the interactions between diversity and \textbf{\textit{Out}}-of-distribution classes.\looseness=-1}    \label{fig:samples-strategy}
\end{figure}

\noindent\textbf{Setup\ \ }
We use ImageNet-100~\cite{imagenet100} as $\mathbb{D}_\text{task}$, and we construct multiple $\mathbb{D}_\text{SSL}$ to evaluate the effects of different data collection strategies and the distribution shift. We first introduce a set $\mathbb{B}$ composed of $65\times 10^3$ images from ImageNet-100 (50\% of the dataset), which we use as a baseline for $\mathbb{D}_\text{SSL}$ with minimum diversity. We denote the 100 classes in ImageNet-100 as \(\mathcal{T}_\text{100}\), and sample $\mathbb{B}$ uniformly including images from all classes. Next, we compare with pretraining on more diverse datasets as $\mathbb{D}_\text{SSL}$, imposing $\mathbb{D}_\text{SSL} = \mathbb{B} \cup \mathbb{A}$ where $\mathbb{A}$ includes $65\times 10^3$ images sampled with one of three strategies. To highlight the effects of the distribution shift, we include in $\mathbb{A}$ either images from \textbf{\textit{In}}-distribution classes, \ie, selecting images from classes overlapping with $\mathcal{T}_\text{100}$, or images from \textbf{\textit{Out}}-of-distribution classes, which do not overlap with $\mathcal{T}_\text{100}$. This is to study the effects of the distribution shift, since we do \textit{not} assume access to downstream classes in real scenarios.
For the \textbf{\textit{Out}}-of-distribution samples, we define $\mathbb{A}$ as (1) random images sampled from the full ImageNet~\cite{imagenet} dataset; (2) images crawled from Flickr, Bing, and DuckDuckGo; or (3) images generated with Stable Diffusion V2.1~\cite{rombach2022high}.
We respectively call these sets
\textcolor{source}{$\mathbb{A}_\text{Source}^\text{\textbf{\textit{Out}}}$},
\textcolor{web}{$\mathbb{A}_\text{Web}^\text{\textbf{\textit{Out}}}$}, and
\textcolor{synthetic}{$\mathbb{A}_\text{Synthetic}^\text{\textbf{\textit{Out}}}$}.
We also define their \textbf{\textit{In}}-distribution counterparts as
\textcolor{source}{$\mathbb{A}_\text{Source}^\text{\textbf{\textit{In}}}$},
\textcolor{web}{$\mathbb{A}_\text{Web}^\text{\textbf{\textit{In}}}$}, and
\textcolor{synthetic}{$\mathbb{A}_\text{Synthetic}^\text{\textbf{\textit{In}}}$}, respectively. Note how $\mathbb{B} \cup \textcolor{source}{\mathbb{A}_{\text{Source}}^\text{\textbf{\textit{In}}}}$ is the full ImageNet-100, coherently with Section~\ref{sec:exp-part1}. Figure~\ref{fig:samples-strategy} shows each collection strategy, for which we provide implementation details in supplementary material. Although many factors (such as the images appearance) impact the distribution shift, using a class prior with any strategy would result in closer distribution with respect to the same strategy \textit{without} class priors~\cite{coleWhenDoesContrastive2022}. We pretrain a ResNet-18 from scratch, with the same settings of Section~\ref{sec:exp-part1}, and \(\mathcal{C}=50\times 10^{6}\). Note that this results in $\mathcal{D} = 1.25 \times 10^{-3}$ for pretraining on $\mathbb{B}$, and $\mathcal{D} = 2.5 \times 10^{-3}$ for pretraining on $\mathbb{B}\cup\mathbb{A}$, introducing a $\mathcal{D}$ difference of a factor of 2.\looseness=-1

\begin{figure*}[t]
    \centering
    \resizebox{\linewidth}{!}{%
    \begin{tabular}{c}
         \includegraphics[width=\linewidth]{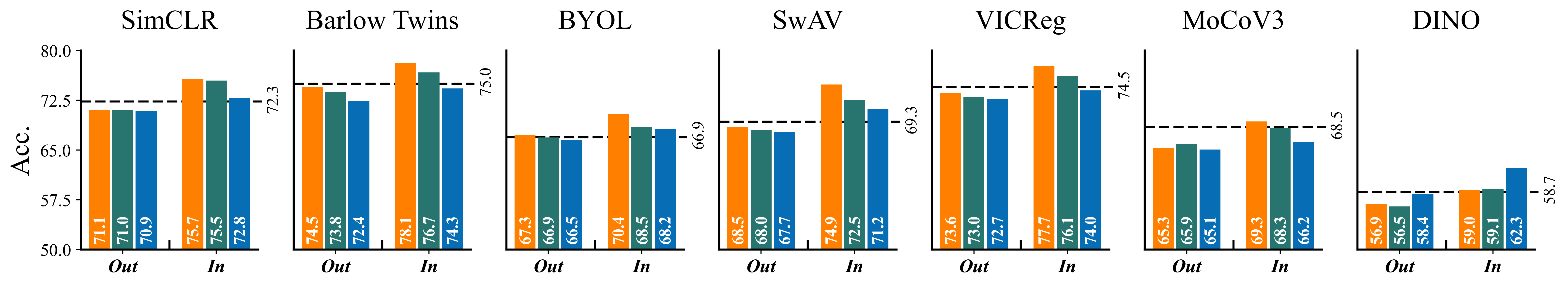} \\
         \includegraphics[width=0.8\linewidth]{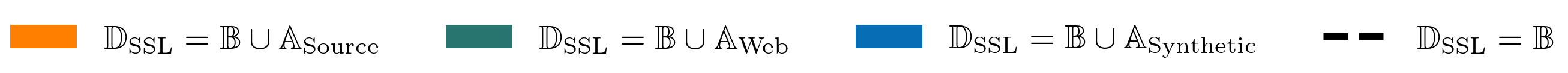}\\
    \end{tabular}%
    }

\caption{\textbf{Effect of Various Data Sources on SSL Pretraining}: We use a baseline set $\mathbb{B}$ (black dashed line), comprising $65\times 10^3$ images from ImageNet-100, for pretraining a ResNet-18 with $\mathcal{C}=50\times10^6$. Augmenting $\mathbb{B}$ with \textbf{\textit{In}}-distribution images enhances performance (above black line), while \textbf{\textit{Out}}-of-distribution augmentations reduce it (below black line). }

    \label{fig:barplots}
\end{figure*}

\noindent\textbf{Results\ \ } We report the linear probing accuracies for pretraining on each $\mathbb{B}\cup\mathbb{A}$ as bars in Figure~\ref{fig:barplots}, showing the $\mathcal{C}$-normalized training on $\mathbb{B}$ as a dashed line. Surprisingly, without class priors (\textbf{\textit{Out}}), including \textcolor{source}{$\mathbb{A}_\text{Source}^\text{\textbf{\textit{Out}}}$}, \textcolor{web}{$\mathbb{A}_\text{Web}^\text{\textbf{\textit{Out}}}$}, and \textcolor{synthetic}{$\mathbb{A}_\text{Synthetic}^\text{\textbf{\textit{Out}}}$} underperforms compared to pretraining on $\mathbb{B}$ only. For instance, while SimCLR on $\mathbb{B}$ scores 72.3\% accuracy, increasing diversity reduces the accuracy by 1\% in all cases. This might appear to conflict with our findings in Section~\ref{sec:exp-part1}, however, the inclusion of \textbf{\textit{Out}} samples leads to \(\mathbb{D}_\text{SSL} \neq \mathbb{D}_\text{task}\), since we sample only classes not included in $\mathbb{D}_\text{task}$. We infer that, with normalized $\mathcal{C}$, increasing $\mathcal{D}$ without distribution priors may negatively impact performance. Conversely, when class priors are available (\textbf{\textit{In}}), increasing pretraining diversity $\mathcal{D}$ improves performance compared to $\mathbb{B}$ pretraining. For instance, pretraining on \textcolor{web}{$\mathbb{A}_\text{Web}^{\text{\textbf{\textit{In}}}}$} performs comparably to augmenting with additional ImageNet samples (\textcolor{source}{$\mathbb{A}_\text{Source}^{\text{\textbf{\textit{In}}}}$}), as in the case of SimCLR where the inclusion of \textcolor{web}{$\mathbb{A}_\text{Web}^{\text{\textbf{\textit{In}}}}$} scores only 0.2\% lower than \textcolor{source}{$\mathbb{A}_\text{Source}^{\text{\textbf{\textit{In}}}}$}. 
Including \textcolor{synthetic}{$\mathbb{A}_\text{Synthetic}^{\text{\textbf{\textit{In}}}}$} data also helps, although more limitedly due to the visual discrepancies between generated and real images. Ultimately, the effectiveness of diversity is linked to the distribution shift. These findings highlight the impact of the distribution shift on computationally-normalized SSL pretraining and help define evaluation practices for this task (see Section~\ref{sec:discussion}).\looseness=-1

\begin{conclusions}[colback=white]
When the distributions of the upstream and downstream tasks differ,  \(\mathbb{D}_\text{SSL} \neq \mathbb{D}_\text{task}\), and in a normalized computation setup, increasing pretraining diversity \(\mathcal{D}\) may harm SSL pretraining performance, reflecting the adverse effects of distribution shift.

\end{conclusions}

\begin{table*}[t]
	\centering
 \caption{\textbf{Performance of ImageNet and YFCC100M SSL Pretraining on Various Downstream Tasks}: We train ResNet-50 and ViT-B/16 under normalized computation ($\mathcal{C}=98\times 10^6$) using SimCLR (left) and MoCoV3 (right) on ImageNet (IN) and YFCC100M (YFCC) with multiple $\mathcal{D}$. Despite the significantly larger $\mathcal{D}$ when trained on YFCC100M, these models cannot offset the effects of the distribution shift and are outperformed by models pretrained on ImageNet in most downstreams.}
 \label{tab:at-scale}
	\setlength{\tabcolsep}{0.008\linewidth}
	\resizebox{0.495\linewidth}{!}{\Large
		\begin{tabular}{cccc|cccccc}
        \multicolumn{10}{c}{\large{\textbf{SimCLR}}}\\
        \toprule
		& \multirow{2}{*}{$\mathbb{D}_\text{SSL}$} & $N$ & $\mathcal{D}$  & \multicolumn{6}{c}{\textbf{Accuracy$\uparrow$}}\\
		& & {$\times 10^6$} & {$\times 10^{-3}$} & IN & Cars & Flow. & Pets & Places & Food\\
		\midrule
		\multirow{5}{*}{\rotatebox{90}{ResNet-50}} & IN & 0.128 & 1.31 & 56.9 & 43.0 & 82.3 & 73.9 & 45.4 & 59.9\\
		& IN & 0.256 & 2.61 & 61.1 & 45.5 & 84.0 & 76.0 & 47.0 & 64.4\\
		  & IN & 0.640 & 6.54 & 63.7 & \textbf{46.8} & 84.2 & 78.8 & 48.3 & 67.2\\
		  & IN & 1.281 & 13.0 & \textbf{64.5} & 46.4 & \textbf{84.6} & \textbf{79.5} & 48.8 & \textbf{68.0}\\
    \cdashlinelr{2-10}
		  & YFCC & 98.17 & 1000 & 57.3 & 37.2 & 76.6 & 58.9 & \textbf{50.1} & 62.1\\
		 \cmidrule{1-10}
		\multirow{5}{*}{\rotatebox{90}{ViT-B/16}} & IN & 0.128 & 1.31 & 54.2 & 37.2 & 81.8 & 70.3 & 44.6 & 64.3\\
		  & IN & 0.256 & 2.61 & 61.3 & 39.5 & 83.8 & 77.7 & 47.2 & 69.4\\
		& IN & 0.640 & 6.54 & 65.5 & \textbf{39.9} & \textbf{84.6} & 80.4 & 49.1 & 73.1\\
		& IN & 1.281 & 13.0 & \textbf{66.7} & 39.5 & 83.7 & \textbf{81.7} & \textbf{50.0} & \textbf{73.6}\\
      \cdashlinelr{2-10}

		  & YFCC & 98.17 & 1000 & 54.5 & 25.0 & 72.5 & 59.7 & 49.5 & 65.4\\
		\bottomrule
        \\[-1.7\medskipamount]
        \multicolumn{10}{c}{$\mathcal{C}=98\times 10^6$}
		\end{tabular}
	}
    \hfill
	\resizebox{0.495\linewidth}{!}{\Large
	\begin{tabular}{cccc|cccccc}
         \multicolumn{10}{c}{\large{\textbf{MoCoV3}}}\\
		\toprule
		& \multirow{2}{*}{$\mathbb{D}_\text{SSL}$} & $N$ & $\mathcal{D}$  & \multicolumn{6}{c}{\textbf{Accuracy$\uparrow$}}\\
		& & {$\times 10^6$} & {$\times 10^{-3}$} & IN & Cars & Flow. & Pets & Places & Food\\
		\midrule
		\multirow{5}{*}{\rotatebox{90}{ResNet-50}} & IN & 0.128 & 1.31 & 58.1 & 40.6 & 81.8 & 76.04 & 45.1 & 63.9\\
		& IN & 0.256 & 2.61 & 62.9 & 45.3 & 85.0 & 79.8 & 47.6 & 68.7\\
		& IN & 0.640 & 6.54 & 65.4 & 48.4 & 86.1 & 81.9 & 49.2 & 71.0\\
		& IN & 1.281 & 13.0 & \textbf{65.9} & \textbf{48.8} & \textbf{86.6} & \textbf{82.6} & 49.5 & \textbf{71.9}\\
        \cdashlinelr{2-10}
		& YFCC & 98.17 & 1000 & 60.4 & 42.2 & 82.6 & 66.3 & \textbf{50.7} & 67.3\\
		\cmidrule{1-10}
		\multirow{5}{*}{\rotatebox{90}{ViT-B/16}} & IN & 0.128 & 1.31 & 57.9 & 33.4 & 82.8 & 78.0 & 46.7 & 67.8\\
		& IN & 0.256 & 2.61 & 63.7 & 35.0 & 85.3 & 82.9 & 48.4 & 71.0\\
		& IN & 0.640 & 6.54 & 67.2 & 39.5 & 85.0 & 85.8 & 49.7 & 72.8\\
		& IN & 1.281 & 13.0 & \textbf{68.8 }& \textbf{41.9} & \textbf{86.5} & \textbf{86.5} & \textbf{50.3} & \textbf{73.8}\\
        \cdashlinelr{2-10}
		& YFCC & 98.17 & 1000 & 57.2 & 25.2 & 70.3 & 43.8 & \textbf{50.3} & 64.0\\
		\bottomrule
        \\[-1.7\medskipamount]
        \multicolumn{10}{c}{$\mathcal{C}=98\times 10^6$}
	\end{tabular}
}
\end{table*}
\subsection{Scaling Pretraining Diversity}\label{sec:exp-part3}

We showed that diversity improves pretraining performance when the training set and downstream task share the same data distribution ($\mathbb{D}_{\text{SSL}} = \mathbb{D}_{\text{task}}$), as discussed in Section~\ref{sec:exp-part1}. This may change when a distribution shift is introduced, as explored in Section~\ref{sec:exp-part2}.
However, it is still unclear from Section~\ref{sec:exp-part2}, whether including a larger number of samples, and thus increasing considerably the pretraining diversity, can compensate for the negative effects of the distribution shift.
To address this, the following section presents larger-scale experiments, employing significantly varied $\mathcal{D}$ values, aiming to explore the interplay between pretraining diversity and different distributions using millions of samples. \\

\noindent\textbf{Setup\ \ } For our large-scale pretraining experiments, we set $\mathbb{D}_\text{SSL}$ to be two datasets of significantly different sizes: ImageNet~\cite{imagenet} and YFCC100M~\cite{thomeeYFCC100MNewData2016}, comprising 1.28 million and 98 million images, respectively. Following  Section~\ref{sec:exp-part1}, we explore multiple \(\mathcal{D}\) values for pretraining. We set \(\mathcal{C}=98\times 10^6\), which corresponds to one epoch on YFCC100M, iterating once through each of its 98 million images to maximize diversity ($\mathcal{D}=1$). Normalizing $\mathcal{C}$ 
(see Section~\ref{sec:exp-part1}), we pretrain on ImageNet for approximately 77 epochs, cumulatively utilizing 98 million samples. Due to the extensive cost of these experiments, we focus on SimCLR and MoCoV3 only. We also employ larger architectures, namely ResNet-50~\cite{heDeepResidualLearning2016} and ViT-B/16~\cite{dosovitskiyImageWorth16x162021}, to leverage the extensive scale of the pretraining datasets.
We also use multiple \(\mathbb{D}_\text{task}\) including ImageNet~\cite{imagenet}, Stanford Cars~\cite{stanfordcars}, Flowers-102~\cite{flowers102}, Oxford-IIIT Pets~\cite{pets}, Places365~\cite{places365}, and Food-101~\cite{food101}.\\

\noindent\textbf{Results\ \ } The results of our large-scale experiments are detailed in Table~\ref{tab:at-scale}. Consistently with findings in Section~\ref{sec:exp-part1}, increasing \(\mathcal{D}\) leads to better pretraining efficacy when \(\mathbb{D}_\text{SSL} = \mathbb{D}_\text{task}\). This is evident when ImageNet is used for both pretraining and downstream classification, reinforcing that our observations hold even at a larger scale. Instead, models pretrained on YFCC100M show substantially lower performance compared to those pretrained on ImageNet, although having much higher $\mathcal{D}$. This highlights the inefficiency of collecting data indiscriminately without considering distribution shifts. To stress this, note how the model pretrained on YFCC100M ($\mathcal{D}=1$) often performs similarly to those pretrained with drastically lower $\mathcal{D}$ on ImageNet ($\mathcal{D}=1.31\times 10^{-3}$). This aligns with our observations in Section~\ref{sec:exp-part2}, emphasizing that distribution differences remain a significant factor even when training with large datasets. However, the YFCC100M-pretrained model outperforms the ImageNet-pretrained model in Places365, suggesting a closer distribution relationship between YFCC100M and Places365. We explore this hypothesis further in Section~\ref{sec:ablations}, where we analyze distribution distances with existing metrics. Ultimately, our analysis highlights that scaling the data is not an effective solution for compensating the distribution shift, when computational resources are normalized. %
 
\begin{conclusions}[colback=white]
Even an extremely large data diversity cannot mitigate the distribution shift under normalized computation. This emphasizes the importance of further research in how effectively SSL pretraining algorithms generalize.
\end{conclusions}

\ifeccv
\vspace{-11px}
\fi
\section{Additional Analysis}
\label{sec:ablations}
\setlength{\fboxsep}{1pt} %
\setlength{\fboxrule}{0pt} %

Previously, we introduced a normalized evaluation to study the relation of $\mathcal{D}$ and data distributions. We showed that, although diversity helps when $\mathbb{D}_\text{SSL}=\mathbb{D}_\text{task}$, vision models are unable to compensate for the distributional differences with normalized costs. We now analyze additional elements supporting our findings.\\

\noindent\textbf{Distribution Distances Using FID \& VisualDNA\ \ }
In Section \ref{sec:exp-part3}, we showed that pretraining on ImageNet typically outperforms YFCC100M on a variety of downstream tasks, with Places365 being the exception. We speculate that the distribution of ImageNet is more aligned with those of the downstream tasks compared to YFCC100M.
To verify this, we evaluate the similarity between the datasets using FID~\cite{fid} and VisualDNA~\cite{ramtoulaVisualDNARepresenting2023} and report results in Table~\ref{tab:visualdna}. With both FID or VisualDNA, the distribution of ImageNet is always closer to the downstream tasks, except for Places365 where YFCC100M is closer. 
This aligns with the lower performance of ImageNet on this dataset only (Table \ref{tab:at-scale}), further suggesting that the performance drop is caused by the distribution shift.\\

\noindent\textbf{Importance of Normalized Computation\ \ } We now study the impact of normalizing computation on the performance of SSL methods. We aim to understand if $\mathcal{C}$ is not normalized across methods, this will lead to misleading conclusions, and thus motivating our normalized computation. In Table~\ref{tab:inconsistencies}, we compare the performance of ResNet-50 and ViT-B/16 models pretrained using MoCoV3 with  \(\mathcal{C} = 98\times 10^6\) (as used in Section~\ref{sec:exp-part3}) against a tripled budget of \(\mathcal{C} = 294\times 10^6\). 
We show that inconsistencies arise when pretraining data diversity $\mathcal{D}$ and computation $\mathcal{C}$ are not fairly compared.
For instance, in scenarios highlighted in \colorbox{red!20}{red}, we notice that a lower pretraining diversity coupled with a higher computational budget can yield better results. We observe that under pretraining diversity of only $\mathcal{D} = 13 \times 10^{-3}$ and a computational budget of $\mathcal{C} = 98 \times 10^{6}$ ResNet-50 only enjoys a 65.9\% accuracy compared against with 69.8\% with a smaller pretraining diversity of $2.17 \times 10^{-3}$ but with a larger computational budget of $294 \times 10^{6}$. This shows that without normalized computation, it could be incorrectly concluded that pretraining diversity does not play a significant role.\\

\begin{table}[t!]
\centering
\begin{minipage}{.48\textwidth}
    \caption{\textbf{Comparing VisualDNA and FID Scores Across Datasets}: We assess the relationship between VisualDNA~\cite{ramtoulaVisualDNARepresenting2023} and FID Score~\cite{fid} for various large-scale $\mathbb{D}_\text{SSL}$ and several downstream tasks. For VisualDNA, activations are extracted as suggested~\cite{ramtoulaVisualDNARepresenting2023} with MUGS~\cite{zhou2022mugs} with ViT-B/16 (MUGS) or DINO v2~\cite{oquab2023dinov2} with ViT-B/16 (DINO v2), while FID activations are obtained via an Inception network~\cite{szegedy2016rethinking}. Consistently, across all metrics, the ranking of dataset distances between $\mathbb{D}_\text{SSL}$ and various $\mathbb{D}_\text{task}$ aligns with the accuracy ranking in Table~\ref{tab:at-scale}. Models exhibiting lower VisualDNA/FID scores benefit more from the diversity in pretraining data for SSL.}
    \label{tab:visualdna}
  \centering
 \resizebox{1\linewidth}{!}{
    \begin{tabular}{cccccccc}
    \toprule
    \multirow{2}{*}{$\mathbb{D}_\text{SSL}$} & \multirow{2}{*}{\textbf{Backbone}} & \multicolumn{6}{c}{\textbf{VisualDNA$\downarrow$}} \\
    & & IN & Cars & Flow. & Pets & Places & Food \\
    \midrule
    IN & MUGS & \textbf{0} & \textbf{11.49} & \textbf{12.46} & \textbf{6.09} & 7.19 & \textbf{9.08}\\
    YFCC & MUGS & 3.72 & 11.57 & 12.71 & 7.93 & \textbf{6.20} & 9.72 \\
    \midrule
    IN & DINO v2 & \textbf{0} & \textbf{7.05} & \textbf{6.95} & \textbf{4.43} & 3.52 & \textbf{7.26} \\
    YFCC & DINO v2 & 2.37 & 7.12 & 7.46 & 5.75 & \textbf{2.74} & 7.90\\ 
    \midrule 
    \end{tabular}
    }
    \resizebox{1\linewidth}{!}{
    \begin{tabular}{cccccccc}
    \toprule
        \multirow{2}{*}{$\mathbb{D}_\text{SSL}$} & \multirow{2}{*}{\textbf{Backbone}} & \multicolumn{6}{c}{\textbf{FID$\downarrow$}}\\
        & & IN & Cars & Flow. & Pets & Places & Food\\ \midrule
    IN & Inception  & \textbf{0} & \textbf{143.4} & \textbf{192.9} & \textbf{88.9} & 64.9 & \textbf{114.9} \\
    YFCC & Inception & 48.1 & 174.3 & 214.8 & 145.8 & \textbf{38.9} & 154.5 \\ \bottomrule
    \end{tabular}}
    \vspace{3px} 
\end{minipage}\hfill%
\begin{minipage}{.48\textwidth}
	\caption{\textbf{Importance of Normalization.} We report the accuracy of MoCoV3 trained on ImageNet (IN) with different data diversity and variable $\mathcal{C}$, for the in-distribution assumption ($\mathbb{D}_\text{SSL}=\mathbb{D}_\text{task}$). For a given network, cells in \colorbox{red!20}{red} highlight inconsistencies where although the model trained with $\mathcal{C}=294\times10^6$ has seen less samples, it outperforms the best model trained with one third of the computational budget ($\mathcal{C}=90\times10^6$), showing the importance of normalization for understanding how models exploit data.}\label{tab:inconsistencies}
  \centering
  	\setlength{\tabcolsep}{0.02\linewidth}

       \resizebox{1\linewidth}{!}{
	\begin{tabular}{ccccc|c}
        \multicolumn{6}{c}{\large{\textbf{MoCoV3}}}\\
		\toprule
        \multirow{2}{*}{\textbf{Network}} & \multirow{2}{*}{$\mathbb{D}_\text{SSL}$} & $\mathcal{C}$ & $N$ & $\mathcal{D}$ & \textbf{Accuracy$\uparrow$}\\
        & & \small{$\times 10^6$} & \small{$\times 10^6$} & \small{$\times 10^{-3}$} &  IN \\
                \midrule
        \multirow{6}{*}{ResNet-50} & IN & $98$ & 0.128 & 1.31 & 58.1 \\ 
		& IN & $98$ & 0.640 & 6.54 & 65.4 \\ 
		& IN & $98$ & 1.281 & 13.0 & \cellcolor{red!20}65.9 \\ 
	   \cdashlinelr{2-6}
        & IN & $294$ & 0.128 & 0.43  & 57.4 \\ 
		& IN & $294$ & 0.640 & 2.17 & \cellcolor{red!20}69.8 \\ 
		& IN & $294$ & 1.281 & 4.35 & 71.4 \\
        \midrule \midrule
        \multirow{6}{*}{ViT-B/16} & IN & $98$ & 0.128 & 1.31  & 57.9 \\ 
		& IN & $98$ & 0.640 & 6.54 & 67.2 \\ 
		& IN & $98$ & 1.281 & 13.0 & \cellcolor{red!20}68.8 \\ 
 \cdashlinelr{2-6}
        & IN & $294$ & 0.128 & 0.43 & 56.9 \\ 
		& IN & $294$ & 0.640 & 2.17 & \cellcolor{red!20}71.9 \\ 
		& IN & $294$ & 1.281 & 4.35 & 74.9 \\
        \bottomrule
	\end{tabular}
	}
\end{minipage}
\vspace{-0.55cm}
\end{table}

\begin{figure}[t]
    \centering
    \resizebox{\linewidth}{!}{%
    \begin{tabular}{c}
    \ifcvpr
    \includegraphics[width=0.6\linewidth]{material/images/saturation/saturation_TN.png}\\
    \includegraphics[width=0.4\linewidth]{material/images/saturation/legend_saturation_TN.png}
    \else
        \includegraphics[width=\linewidth]{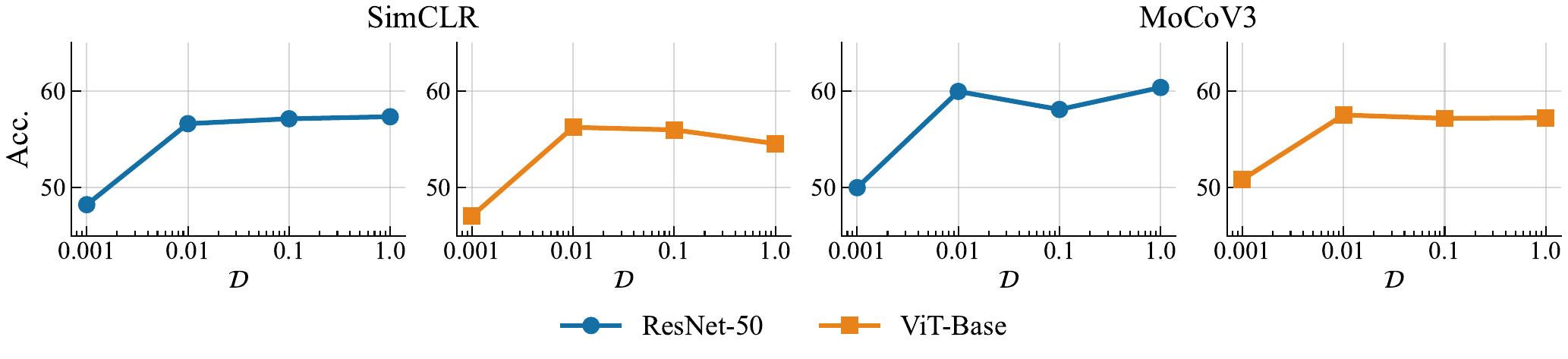}
    \fi
         \end{tabular}%
    }
    \\[-1.1ex]
    \caption{\textbf{Data Diversity Impact on YFCC100M Pretraining Performance}: Pretraining ($\mathcal{C}=98\times10^6$) of networks with $\mathbb{D}_\text{SSL}=\text{YFCC100M}$ and $\mathbb{D}_\text{task}=\text{ImageNet}$ for several dataset subsets. In the presence of a distribution shift, performance tends to saturate and does not benefit from additional data diversity.}
    \label{fig:saturation}\vspace{-0.25cm}
\end{figure}

\noindent\textbf{Model Saturation\ \ } 
We aim to study the pretraining trend on YFCC100M, and if adding more samples could compensate the distribution shift with ImageNet. Hence, we extend the YFCC100M experiments from Section~\ref{sec:exp-part3}, examining various subsets---specifically 0.1\%, 1\%, 10\%, and 100\% of the dataset. Again, we normalize the computational budget to \(\mathcal{C} = 98\times 10^6\), equivalent to one epoch on YFCC100M. The results of linear probing models pretrained using SimCLR and MoCoV3 with ResNet-50 and ViT-B/16 on ImageNet are shown in Figure~\ref{fig:saturation}. Interestingly, we obtain a performance plateau. This observation points to a saturation point in performance, showing that simply increasing the dataset further would not bridge the gap between the $\mathbb{D}_\text{SSL}$ and $\mathbb{D}_\text{task}$. Hence, arbitrarily increasing pretraining diversity $\mathcal{D}$ is not sufficient to bridge the distribution shift problem. Here \(\mathbb{D}_{\text{SSL}} \neq \mathbb{D}_{\text{task}}\), hence still aligning with our findings in Section ~\ref{sec:exp-part1}.

\noindent\textbf{Label Quantity\ \ } A question that arises in our setting is whether having a higher pretraining diversity leads to requiring less labeled samples during linear probing. 
In this section, we focus on understanding the impact of increased $\mathcal{D}$
on the number of labeled samples required for effective linear probing in downstream tasks.
We use the same trained models from Section~\ref{sec:exp-part3}, \ie, SimCLR and MoCoV3, with ResNet-50 and ViT-B/16 architectures. %
In this setup, we set $\mathbb{D}_\text{task}=\text{ImageNet}$, with two upstream dataset scenarios: ImageNet (where \(\mathbb{D}_{\text{SSL}} = \mathbb{D}_{\text{task}}\)) and YFCC100M (where \(\mathbb{D}_{\text{SSL}} \neq \mathbb{D}_{\text{task}}\)). Our experiments %
are summarized in Table~\ref{tab:label_quantities}. We note that with ViT-B/16, if ImageNet is used for pretraining, linear probing with just 5\% labeled samples can surpass the performance achieved using 100\% labeled data when YFCC100M serves as $\mathbb{D}_\text{SSL}$. This also implies that when \(\mathbb{D}_{\text{SSL}} \neq \mathbb{D}_{\text{task}}\), a higher quantity of labeled data is necessary to attain competitive performance, and increasing $\mathcal{D}$ %
does not bring considerable benefits on label quantity under distribution shift. This implies that our findings in Section~\ref{sec:exp-part3} hold regardless of the linear probing labeled set. We note that in scenarios where \(\mathbb{D}_{\text{SSL}} = \mathbb{D}_{\text{task}}\), using only 50\% of the labeled data can achieve similar performance as utilizing the full 100\% of labeled samples, implying that increasing $\mathcal{D}$ leads to reduced label requirement efforts for downstream tasks.\looseness=-1

\begin{table}[t]
 \caption{\textbf{Evaluating Network Accuracy With Varied Label Quantity}: We evaluate the accuracy of networks trained on ImageNet (IN) and YFCC100M (YFCC) with different labeling percentages of $\mathbb{D}_\text{task}=\text{ImageNet}$. The increased diversity still does not compensate for the distribution shift. However, for in-distribution data, one can get away with fewer labels with more diverse pretraining data.} \label{tab:label_quantities}
\small
	\centering
	\setlength{\tabcolsep}{0.01\linewidth}
\setlength{\aboverulesep}{0pt}
\setlength{\belowrulesep}{0pt}
	\resizebox{0.495\linewidth}{!}{
	\begin{tabular}{cccc|cccc}
        \multicolumn{8}{c}{{\textbf{SimCLR}}}\\
		\toprule
		\multirow{2}{*}{\textbf{Network}} &  \multirow{2}{*}{$\mathbb{D}_\text{SSL}$} & $N$ & $\mathcal{D}$  & \multicolumn{4}{c}{\textbf{Accuracy$\uparrow$}}\\
		& & \small{$\times 10^6$} & \small{$\times 10^{-3}$} & 5\% & 10\% & 50\% & 100\% \\
		\midrule
		\multirow{2}{*}{ResNet-50}  & \small IN & 1.28 & 1.31 & \textbf{50.3} & \textbf{54.2} & \textbf{61.7} & \textbf{64.5} \\
		& \small YFCC & 98.2 & 1000 & 39.4 & 44.3 & 54.0 & 57.3 \\
		\midrule
		\multirow{2}{*}{ViT-B/16}  & \small IN & 1.28 & 1.31 & \textbf{55.2} & \textbf{59.4} & \textbf{65.6} & \textbf{66.7} \\
		& \small YFCC & 98.2 & 1000 & 40.7 & 45.7 & 53.3 & 54.5 \\
		\bottomrule
        \\[-1.5\medskipamount]
        \multicolumn{8}{c}{$\mathcal{C}=98\times10^6$}
	\end{tabular}
	}
	\resizebox{0.495\linewidth}{!}{
     \begin{tabular}{cccc|cccc}
        \multicolumn{8}{c}{{\textbf{MoCoV3}}}\\
		\toprule
		\multirow{2}{*}{\textbf{Network}} &  \multirow{2}{*}{$\mathbb{D}_\text{SSL}$} & $N$ & $\mathcal{D}$  & \multicolumn{4}{c}{\textbf{Accuracy$\uparrow$}}\\
		& & \small{$\times 10^6$} & \small{$\times 10^{-3}$} & 5\% & 10\% & 50\% & 100\% \\
		\midrule
		\multirow{2}{*}{ResNet-50}  & \small IN & 1.28 & 1.31 & \textbf{52.1} & \textbf{56.2} & \textbf{63.8} & \textbf{65.9} \\
		& \small YFCC & 98.2 & 1000 & 42.8 & 48.2 & 57.7 & 60.4 \\
		\midrule
		\multirow{2}{*}{ViT-B/16} & \small IN & 1.28 & 1.31 & \textbf{59.3} & \textbf{63.0} & \textbf{68.3}& \textbf{68.8} \\
		& \small YFCC & 98.2 & 1000 & 43.9 & 49.0 & 56.0 & 57.2 \\
		\bottomrule
        \\[-1.5\medskipamount]
        \multicolumn{8}{c}{$\mathcal{C}=98\times10^6$}
 \end{tabular}}
\end{table}

\section{Discussion}\label{sec:discussion}

\noindent\textbf{Main Conclusions\ \ } Our set of experiments leads to Insight (3) in Section~\ref{sec:exp-part3}, revealing that with normalized computational costs, SSL pretrainings with large diversity cannot compensate for the distribution shift. This is surprising, since the variety of information that SSL algorithms could benefit from during training is much higher in large generic datasets than in small ones.
Since our evaluation is cost-normalized, (3) also shows that SSL strategies \textit{are not efficiently exploiting the pretraining diversity on large datasets} for representation extraction. This inefficiency implies a wide margin for improvement of generalization of SSL models, making better use of the computational power used for training. The role of existing models must also be discussed in this context. Following Insights (1) and (2), respectively in Sections~\ref{sec:exp-part1} and~\ref{sec:exp-part2}, we have studied how in-distribution and out-of-distribution data impact performance in a controlled scenario. We argue that this should be taken into account in the evaluation of the performance of SSL models. 
Indeed, training at scale may enlarge the in-distribution space, including classes of the downstream tasks in the training set. In recent literature, this is a design choice to maximize performance on popular benchmarks~\cite{oquab2023dinov2}. While this allows for achieving impressive results in many tasks, we stress that this does not permit a fair evaluation. Now, we summarize practical takeaways.\looseness=-1

\noindent\textbf{Training Takeaways\ \ } Coherently with our findings in Section~\ref{sec:exp-part1} and similarly to prior art~\cite{kotar2021contrasting, coleWhenDoesContrastive2022, goyal2019scaling}, we find that aligned distributions benefit performance, in particular increasing $\mathbb{D}_\text{SSL}$ diversity helps, as long as the distribution of the new samples match those of the downstream task data. %
Differently from the state-of-the-art, we demonstrate that this holds also in a computationally-normalized setting, implying that collecting large scale in-distribution data matching the downstream task is be an effective approach for improving SSL. Hence, for practical applications distribution priors should be used, if available. On the contrary, for a fair evaluation of models, this should not be the case, as specified below.

\noindent\textbf{Evaluation Takeaways\ \ } %
Our analysis reveals that to permit a fair evaluation of SSL methods, computationally normalized tests are necessary to avoid inconsistencies, as shown in Section~\ref{sec:ablations}. Moreover, it is crucial to identify out-of-distribution downstream tasks for a correct evaluation of generalization performance. By evaluating only on $\mathbb{D}_\text{task}$ with a low distribution shift, there is a risk of reporting inflated metrics, not representative of a real gain in generalization. This is important, since new SSL approaches may be reporting higher downstream performance when pretrained on a different dataset. We relate this to Sections \ref{sec:experiments} and \ref{sec:ablations}, where we show that increasing the computation and the in-distribution data, respectively, can improve performance. Ultimately, wrong practices may result in incorrectly concluding that an SSL algorithm is performing better.\looseness=-1

\noindent\textbf{Differences With Language Models\ \ } 
In Section~\ref{sec:exp-part3} we showed that even very diverse datasets (YFCC100M), fall short in satisfactory generalization performance. In addition to further exploration into generalization for SSL pretraining, this open doors to investigating why language models enjoy enhanced generalization when trained on highly diverse data compared to vision models~\cite{wei2022emergent}.\looseness=-1

\section*{Acknowledgements}

This work was supported by SDAIA-KAUST Center of Excellence in Data Science, Artificial Intelligence (SDAIA-KAUST AI). Fabio Pizzati is financed by KAUST (Grant DFR07910). Philip H.S. Torr thanks the Royal Academy of Engineering for their support. This work is supported by a UKRI grant Turing AI Fellowship (EP/W002981/1).

\bibliographystyle{splncs04}
\bibliography{main}

\clearpage
\onecolumn

\setcounter{section}{0}
\definecolor{cvprblue}{rgb}{0.21,0.49,0.74}

\newcommand{\mainref}[1]{\hypersetup{linkcolor=red}\textcolor{red}{\ref{#1}}\hypersetup{linkcolor=red}}
\renewcommand\thesection{\Alph{section}}
\renewcommand\thesubsection{\Alph{section}.\arabic{subsection}}

\MakeTitle{On Pretraining Data Diversity\\ for  
Self-supervised Learning\\
\textit{Supplementary Material
}}{}{}

\noindent In this supplementary material, we present additional experiments and insights on our findings presented in the main paper. First, we further develop the inconsistencies found within an incorrectly-normalized framework (Section~\ref{supp:compute}). Then, we present rerun some of the experiments from Section~\mainref{sec:exp-part1} on multiple seeds to understand the statistical significance of our findings (Section ~\ref{sec:stat_sig}). After, we propose different settings for our analysis at scale (Section~\ref{supp:alternative-settings}). We extend our analysis on label quantity in Section~\ref{supp:labels}. Finally, we introduce additional details about our settings and implementations (Section~\ref{supp:details}). For reproducibility of our results, we will share on GitHub our codebase, along the fine-tuned parameters and the data ordering files.

\section{Importance of Normalization of Computation}\label{supp:compute}

In this section, we aim to complement our experiment in Section~\mainref{sec:ablations}, providing further proof highlighting the importance of having a normalized computational budget when evaluating the performance of SSL methods. In the following experiments, we show that if computation is not normalized properly, one might fall into unfair comparisons and misleading conclusions.

\subsection{Increasing Total Computation}

In our first setup, we pretrain SimCLR~\cite{simclr} and MoCoV3~\cite{mocov3} on Tiny ImageNet with various $\mathcal{D}$, over a range of increasing amounts of budget $\mathcal{C}$. We assume $\mathbb{D}_\text{SSL}=\mathbb{D}_\text{task}$.
We take the same subsets of Tiny ImageNet as in Section~\mainref{sec:exp-part1}, which consist of 10\%, 50\% and 100\% of the training data.
We vary $\mathcal{C}$ from $5\times10^6$ to $100\times10^6$, and we measure the accuracy of the pretrained models on the $\mathbb{D}_\text{task}$ test set, results of which are shown in Table~\ref{tab:increase_compute}.
Note that we refer to models in this section using the data diversity $N$ instead of pretraining diversity $\mathcal{D}$, as each cell in Table~\ref{tab:increase_compute} has a different $\mathcal{D}$.
Following our previous experiments, we argue that comparison between different diversities only hold as long as the computation is normalized.
This implies that comparisons only hold within any given $\mathcal{C}$.
In agreement with our prior findings, the third row with the highest diversity always outperforms the lower pretraining diversity models on in-domain evaluation, for both SimCLR and MoCoV3.
However, only when we compare between different columns, \ie different amounts of computation, we may observe that models pretrained with lower diversity may outperform higher diversity models.
For example, for both SimCLR and MoCoV3, the models pretrained with $N=50 \times 10^3$ and $\mathcal{C}=10 \times 10^6$ outperform the models with higher data diversities $N=100 \times 10^3$ but less computation $\mathcal{C}=5 \times 10^6$.
As a result, we see that models with lower pretraining diversity can still outperform models with higher diversity, given that more computation is used. This highlights the importance of normalizing computation costs when evaluating the effects of diversity.

\begin{table*}[t]
\ifeccv
\caption{\textbf{Pretraining Diversity With Increasing Computational Budget}: We show for both SimCLR (left) and MoCoV3 (right) that increasing pretraining diversity always leads to better in-domain downstream accuracies, given that computation is normalized, \ie, comparisons hold within the columns of the tables. Comparing models between different columns may lead to inconsistencies, where lower diversity models with more computation obtain higher results than higher diversity models with less computation.}
\label{tab:increase_compute}
\fi
	\centering
	\begin{subtable}{0.48\linewidth}
 \setlength\tabcolsep{0.25em} %
\setlength{\aboverulesep}{0pt}
\setlength{\belowrulesep}{0pt}\setlength\tabcolsep{0.25em}
	\resizebox{\linewidth}{!}{%
		\begin{tabular}{cC{1.2cm}|cC{1.2cm}C{1.2cm}C{1.2cm}C{1.2cm}C{1.2cm}}
        \multicolumn{8}{c}{{\textbf{SimCLR}}}\\
		\toprule
		& $N$ && \multicolumn{5}{c}{$\mathcal{C}$\small $\ \ (\times 10^6)$} \\
        & \small $(\times 10^3)$  &&  5 & 10 & 25 & 50 & 100 \\
		\midrule
		& 10  && 36.92&36.63&36.30&36.91&35.03\\
		& 50 && 40.76&44.30&47.69&48.77&48.91 \\
		& 100 && \textbf{41.43}&\textbf{44.76}&\textbf{49.32}&\textbf{49.83}&\textbf{51.62}\\
		\bottomrule
	\end{tabular}
	}
\end{subtable}	
\hfill
\begin{subtable}{0.48\linewidth}
\setlength\tabcolsep{0.25em} %
\setlength{\aboverulesep}{0pt}
\setlength{\belowrulesep}{0pt}\setlength\tabcolsep{0.25em}
	\resizebox{\linewidth}{!}{%
		\begin{tabular}{cC{1.2cm}|cC{1.2cm}C{1.2cm}C{1.2cm}C{1.2cm}C{1.2cm}}
        \multicolumn{8}{c}{{\textbf{MoCoV3}}}\\
		\toprule
		& $N$ && \multicolumn{5}{c}{$\mathcal{C}$\small $\ \ (\times 10^6)$} \\
        & \small $(\times 10^3)$  &&  5 & 10 & 25 & 50 & 100 \\
		\midrule
		& 10  && 39.78&41.82&40.06&36.56&28.92 \\
		& 50 && 39.88&43.42&46.68&46.45&48.14 \\
		& 100 && \textbf{40.35}&\textbf{44.03}&\textbf{47.63}&\textbf{48.58}&\textbf{50.71} \\
		\bottomrule
	\end{tabular}
	}
\end{subtable}
\ifcvpr
\caption{\textbf{Pretraining Diversity With Increasing Computational Budget}: We show for both SimCLR (left) and MoCoV3 (right) that increasing pretraining diversity always leads to better in-domain downstream accuracies, given that computation is normalized, \ie, comparisons hold within the columns of the tables. Comparing models between different columns may lead to inconsistencies, where lower diversity models with more computation obtain higher results than higher diversity models with less computation.}
\label{tab:increase_compute}
\fi

\end{table*}

\subsection{Epoch-based normalization}

In Section \mainref{sec:exp-part2}, we adhered to a fixed computational budget of \(\mathcal{C}=50 \times 10^6\), pretraining models on \(\mathbb{D}_{\text{SSL}} = \mathbb{B}\) for 800 epochs, and on \(\mathbb{B} \cup \mathbb{A}\) for 400 epochs, considering that the latter dataset was twice the size. We further demonstrate the importance of a computationally-normalized evaluation by exposing the inconsistencies of an alternative epoch-based normalization, hence in which networks are trained for 400 epochs, regardless of the dataset size.

\begin{figure}[t]
    \centering
    \includegraphics[width=\linewidth]{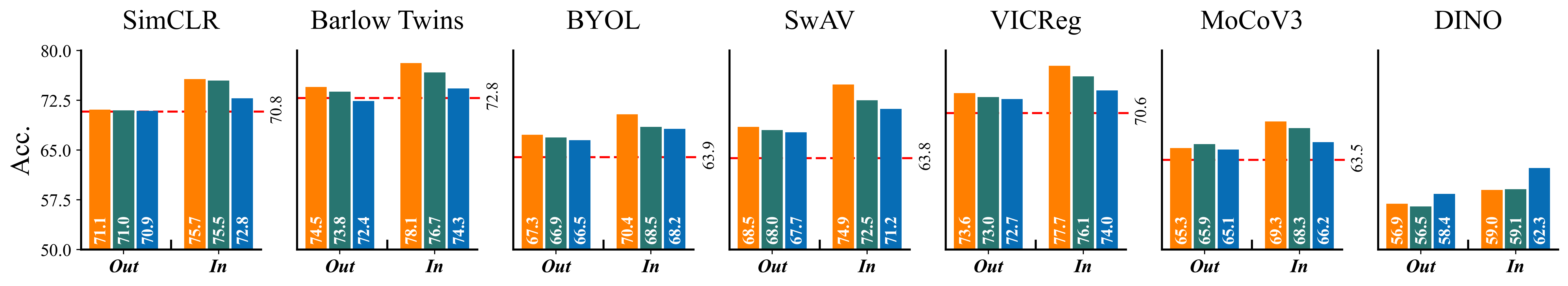}
\caption{\textbf{Impact of Epoch Normalization on SSL Pretraining Performance}: This figure contrasts an epoch-normalized baseline (red line) with the trained methods in the main paper, Figure~\ref{fig:barplots}. Under epoch normalization, we notice contrasting findings, \ie more diverse trainings, irrespective of their origin (source, web, or synthetic) and label distribution (in or out-of-distribution), consistently enhances performance. This is an unfair comparison due to the greater costs of each augmented pretraining if epochs are normalized. This illustrates how alternative normalization can lead to wrong conclusions compared to compute normalization. DINO $\mathbb{B}$ epoch-normalized baseline is shown in text only (Acc. \textcolor{red}{41.14}) for ease of visualization.}
    \label{fig:red_line}
\end{figure}

We propose this alternative scenario in Figure~\ref{fig:red_line}, where the compute-normalized baseline (the black dashed line in Figure \mainref{fig:barplots}) is replaced with an epoch-normalized baseline (indicated by the red dashed line), obtained by pretraining for 400 epochs on $\mathbb{B}$. Here, we observe that augmenting with additional samples consistently enhances performance, irrespective of the data augmentation technique used and whether the sample labels are in or out of distribution. This finding does not align with the insights from Section \mainref{sec:exp-part2}, and we highlight that it does not take into account the difference in costs for training models for the same number of epochs, but on a dataset twice the size. Hence, this constitutes an unfair comparison that may lead to incorrect conclusions, advocating for the effectiveness of our computational-based normalization.

\section{Alternative Settings}\label{supp:alternative-settings}

\subsection{Non-Contrastive Methods}
\label{sec:bt_scale}

For large-scale experiments in section \mainref{sec:exp-part3} we only considered SimCLR~\cite{simclr} and MoCoV3~\cite{mocov3}, both of which are contrastive SSL methods.
Here we show that the results are consistent for the non-contrastive methods Barlow Twins~\cite{barlow} and BYOL~\cite{byol}.
We highlight that these experiments are computationally intensive, hence we explore a reduced setting with a single backbone and lower $\mathcal{D}$ variability.
We pretrained a ResNet-50 backbone using Barlow Twins and BYOL on ImageNet and the same subsets from Section~\mainref{sec:exp-part3}, as well as on the full YFCC100M dataset, ensuring that the total compute is equal to a single epoch on YFCC100M, i.e. $\mathcal{C}=98\times 10^6$.
Again, we show linear evaluation on multiple downstream datasets including ImageNet~\cite{imagenet}, Stanford Cars~\cite{stanfordcars}, Flowers-102~\cite{flowers102}, Oxford-IIIT Pets~\cite{pets}, Places365~\cite{places365}, and Food-101~\cite{food101} in Table~\ref{tab:additional_methods}.
In accordance with Section~\mainref{sec:exp-part3}, we observe that pretraining on higher diversities leads to improved downstream accuracies when $\mathbb{D}_{\textrm{SSL}} = \mathbb{D}_{\textrm{task}}$, i.e. pretraining and evaluating on ImageNet.
Also, the highest pretraining diversity in ImageNet leads to the best results for all downstream datasets, except for Places365, for which pretraining YFCC100M performs best, for which we refer again to distribution distances analysis in Section~\mainref{sec:ablations}.
Again for these methods, the maximum diversity of YFCC100M is not enough to diminish the effects of the domain gap between the pretraining data and the datasets other than Places365.

\begin{table*}[t]
\ifeccv
	\caption{\textbf{Non-contrastive pretraining}: We explore two more pretraining methods, namely, Barlow Twins and BYOL, for our large-scale pretraining experiments. Again, the budget is set to $\mathcal{C}= 98\times 10^6$, we find that our earlier conclusions still hold here: (1) ImageNet pretraining outperforms YFCC100M (YFCC) pretraining except for Places365 due to the distribution shift (2) Increased pretraining diversity $\mathcal{D}$ generally correlates with improved downstream performance with the exception. Those findings are consistent for both Barlow Twins and BYOL.}
    \label{tab:additional_methods}
\fi
	\centering
	\begin{subtable}{0.48\linewidth}
		\setlength\tabcolsep{0.25em} %
	\resizebox{\linewidth}{!}{%
		\begin{tabular}{cccc|cccccc}
        \multicolumn{10}{c}{\large{\textbf{Barlow Twins}}}\\
        \toprule
		& \multirow{2}{*}{$\mathbb{D}_\text{SSL}$} & $N$ & $\mathcal{D}$  & \multicolumn{6}{c}{\textbf{Accuracy$\uparrow$}}\\
		& & $(\times10^6)$ & $(\times10^{-3})$ & ImageNet & Cars & Flow. & Pets & Places & Food\\
		\midrule
		  & ImageNet & 0.128 & 1.31 & 57.17 & 51.51 & 85.84 & 75.81 & 45.87 & 63.72\\
		  & ImageNet & 1.281 & 13.0 & \textbf{65.40} & \textbf{59.56} & 89.16 & \textbf{83.89} & 49.19 & \textbf{70.42} \\
    \cdashlinelr{2-10}
		  & YFCC & 98.17 & 1000 & 57.85 & 44.76 & 83.46 & 67.21 & \textbf{50.01} & 65.20 \\
		\bottomrule
        \\[-1.5\medskipamount]
        \multicolumn{10}{c}{$\mathcal{C}=98\times 10^6$}
		\end{tabular}
	}
\end{subtable}	
\hfill
\begin{subtable}{0.48\linewidth}
\setlength\tabcolsep{0.25em} %
\resizebox{\linewidth}{!}{%
	\begin{tabular}{cccc|cccccc}
        \multicolumn{10}{c}{\large{\textbf{BYOL}}}\\
        \toprule
		& \multirow{2}{*}{$\mathbb{D}_\text{SSL}$} & $N$ & $\mathcal{D}$  & \multicolumn{6}{c}{\textbf{Accuracy$\uparrow$}}\\
		& & $(\times10^6)$ & $(\times10^{-3})$ & ImageNet & Cars & Flow. & Pets & Places & Food\\
		\midrule
		  & ImageNet & 0.128 & 1.31 & 61.82 & 46.62 & 85.84 & 80.28 & 46.91 & 67.01\\
		  & ImageNet & 1.281 & 13.0 & \textbf{68.39} & \textbf{52.51} & \textbf{88.77} & \textbf{84.75} & 49.96 & \textbf{73.52}\\
    \cdashlinelr{2-10}
		  & YFCC & 98.17 & 1000 & 60.73 & 42.78 & 84.38 & 68.63 & \textbf{50.92} & 68.63 \\
		\bottomrule
        \\[-1.5\medskipamount]
        \multicolumn{10}{c}{$\mathcal{C}=98\times 10^6$}
		\end{tabular}
}

\end{subtable}
\ifcvpr
	\caption{\textbf{Non-contrastive pretraining}: We explore two more pretraining methods, namely, Barlow Twins and BYOL, for our large-scale pretraining experiments. Again, the budget is set to $\mathcal{C}= 98\times 10^6$, we find that our earlier conclusions still hold here: (1) ImageNet pretraining outperforms YFCC100M (YFCC) pretraining except for Places365 due to the distribution shift (2) Increased pretraining diversity $\mathcal{D}$ generally correlates with improved downstream performance with the exception. Those findings are consistent for both Barlow Twins and BYOL.}
    \label{tab:additional_methods}
\fi

\end{table*}

\subsection{Different Architectures}

We investigate how pretraining diversity interacts with varying backbone architecture sizes, as well as the total computational budget. With this, we aim to highlight how different models react to pretraining diversity. To benchmark the interaction of these factors, we focus on MoCoV3 and pretrain and evaluate on ImageNet using $\mathcal{C} = 98\times 10^6$ and the tripled amount $\mathcal{C} = 294\times 10^6$.
We use two different architecture sizes for ViT backbones as well as ResNet backbones: ViT-Small/16 paired with ViT-Base/16 and ResNet-50 paired with ResNet-101.

Results are shown in Table~\ref{tab:larger_architectures}, and the first observation we make, is that for any combination of architecture size and total amount of computation, the model pretrained with largest amount of pretraining diversity $\mathcal{D}=13.0\times 10^{-3}$ always has the highest in-domain downstream performance.
Increasing pretraining diversity thus remains a reliable method to improve the quality of learned representations, regardless of the architecture size.
Secondly, we see that for every diversity value, regardless of the backbone type or the amount of computation, an increase in backbone size, \ie ResNet-50 to ResNet-101 or ViT-S/16 to ViT-B/16, leads to an increased performance.
It is thus again of importance to only compare models with different pretraining diversities for fixed model size, as we did with fixed computational budget.

Finally, keep in mind that the larger architectures require more computation, which is not incorporated in $\mathcal{C}$ as this term only describes the number of images that are seen during pretraining.
\subsubsection{A note on MAE} Masked Autoencoders (MAE)~\cite{he2022masked} is a Transformer-specific pretraining method based on an autoregressive loss. This differs considerably from what has been presented in Section~\mainref{sec:prelims}, and it has significant impact on the components needed for our normalized analysis. Indeed, for a $\mathcal{C}=98\times10^6$ budget, MAE is far from providing optimal performance~\cite{he2022masked}, making comparisons unfair without incurring in unsubstainable costs. Also, the reconstruction task used for supervision extracts features requiring a deep decoder for best performance in linear probing~\cite{he2022masked}, and it results in considerably better performance with full finetuning exclusively. We will analyze pretraining diversity effects for MAE in a future work.

\subsection{Convergence insights}

The convergence of models trained on YFCC and Imagenet leading to our Insight 3 must be further discussed (see main paper, Section~\mainref{sec:exp-part3}). One may argue that although $\mathcal{C}=98\times10^6$ maximizes pretraining diversity on YFCC100M, this may not enough for making trained models fully converge. First, we highlight how relevant literature sets similar training budget $\mathcal{C} = 100 \times 10^6$ requirements for drawing reliable conclusions~\cite{simclr}. Secondly, we stress how bringing to convergence both models pretrained on Imagenet and YFCC100M would inevitably result in a different sizing of the computational budget, preventing a fair evaluation. Alternatively, increasing the computational budget for a complete convergence of both settings would inevitably lead to the overfitting of the model trained on Imagenet. This may lead to misleading results, since the overfitting-related loss of performance could lead to wrong conclusions related to the distribution shift impact. Instead, our setup guarantees a reliable evaluation, by preventing overfitting while training enough for a reasonable representation extraction. Moreover, we relate to relevant literature highlighting the importance of single-epoch training for representation extractors~\cite{tong2023emp}.

\begin{table}[t]
\centering
\begin{tabular}{@{}lcccccccc@{}}
\toprule
\multicolumn{9}{c}{\textbf{MoCoV3 - Resnet50}} \\
\midrule
\multirow{2}{*}{$\mathbb{D}_\text{SSL}$} & {$N$} & $\mathcal{D}$ & \multirow{2}{*}{IN} & \multirow{2}{*}{Cars} & \multirow{2}{*}{Flow.} & \multirow{2}{*}{Pets} & \multirow{2}{*}{Places} & \multirow{2}{*}{Food} \\
& $\times 10^6$ & $\times 10^{-3}$ & & & & & \\
\midrule
IN & 1.281 & 4.33 & \textbf{71.41} & 60.37 & 91.28 & 87.15 & 50.80 & 74.70 \\
YFCC & 98.17 & 333 & 62.75 & 43.49 & 82.72 & 68.85 & \textbf{51.99} & 68.63 \\
\bottomrule
\multicolumn{9}{@{}c@{}}{\footnotesize$\mathcal{C} = 294 \times 10^6$} \\
\end{tabular}
\caption{\textbf{Effects of more computation.} We triple the budget $\mathcal{C}$ used in Section 5.3 of the main paper, bringing MoCoV3 to convergence on YFCC. Our insights remain unchanged.}
\label{tab:more_compute}
\end{table}

\subsection{More computation for YFCC trainings}

In Table \ref{tab:more_compute} we show the results of running MoCo V3 ResNet50 for both YFCC and ImageNet datasets with $3 \times$ the computational budget with respect to $\mathcal{C}$ used in Section 5.3 of the main paper. We run a limited set of experiments due to their cost. Importantly, this allows the trained network to converge on YFCC. Our results show that the conclusions are consistent with the findings in Section \ref{sec:exp-part3} where ImageNet as an upstream consistently outperforms YFCC and a higher diversity does not bridge the domain gap between the upstream and downstream.

\section{Statistical Significance}\label{sec:stat_sig}

In this subsection we revisit the experiments from Section \ref{sec:exp-part1} and run CIFAR100 experiments for three difference seeds. The mean and standard deviation of the runs is provided in Table \ref{tab:statistics} where we find that the conclusions still hold where we observe again that when the upstream and downstream distributions are the same, a higher data diversity leads to better downstream accuracy.

\begin{table}[h]
\centering
    \setlength{\aboverulesep}{0pt}
    \setlength{\belowrulesep}{0pt}
    \setlength\tabcolsep{0.18em} %
    \resizebox{\linewidth}{!}{%
        \begin{tabular}{cc|cC{2cm}C{2cm}C{2cm}C{2cm}C{2cm}C{2cm}C{2cm}}
        \toprule
        $N$ & $\mathcal{D}$ & \multicolumn{8}{c}{\textbf{Accuracy$\uparrow$}} \\
        \scriptsize $\times 10^3$ & \scriptsize $\times 10^{-3}$ &&  \small{SimCLR} & \small{B.T.} & \small{BYOL} & \small{SwAV} & \small{VICR} & \small{\mbox{MoCo3}} & \small{DINO}\\
        \midrule
        5  & 0.1 && 38.1 ± 0.2 & 44.8 ± 0.4 & 40.6 ± 0.4 & 14.2 ± 0.1 & 43.6 ± 0.4 & 12.0 ± 0.5 & 29.9 ± 12.2 \\
        25 & 0.5 && 51.1 ± 0.8 & 54.1 ± 0.2 & 51.2 ± 0.1 & 45.7 ± 0.2 & 49.8 ± 0.5 & 51.7 ± 0.1 & 39.8 ± 3.4 \\
        50 & 1.0 && 58.5 ± 0.1 & 58.7 ± 0.3 & 58.1 ± 0.2 & 56.4 ± 0.1 & 55.5 ± 0.3 & 55.2 ± 0.3 & 42.0 ± 0.9 \\
        \bottomrule
        \\[-1.5\medskipamount]
        \multicolumn{10}{c}{$\mathcal{C}=50\times 10^6$}
    \end{tabular}
    }
\caption{\textbf{Statistical Significance for CIFAR100 Experiments.} We investigate the effect of changing the seed on the conclusions of Section \ref{sec:exp-part1} for CIFAR100. Averaging over three seeds, we find that the conclusions remain unchange that when the upstream and downstream have the same distribution, a higher data diversity leads consistently to better results.}\label{tab:statistics}

\end{table}

\begin{table*}[t]
\ifeccv
	\caption{\textbf{Pretraining Diversity With Different Architectures Sizes}: We investigate how pretraining diversity, total computation budget and model architecture size interact for MoCoV3 when pretraining and evaluating on ImageNet. Regardless of $\mathcal{C}$ and the architecture choice, increasing pretraining diversity remains a reliable method to improve downstream results. Further, increasing model size also seems to consistently lead to better learned representations. Again, comparing pretraining diversity values only holds when the model architecture and $\mathcal{C}$ are fixed.}
    \label{tab:larger_architectures}
\fi
	\centering
    \setlength\tabcolsep{0.25em} %
	\resizebox{0.7\linewidth}{!}{%
		\begin{tabular}{cccc|C{1.7cm}C{1.7cm}|C{1.7cm}C{1.7cm}}
        \multicolumn{8}{c}{{\textbf{MoCoV3}}}\\
		\toprule
         \multirow{2}{*}{$\mathbb{D}_\text{SSL}$} & $\mathcal{C}$ & $N$ & $\mathcal{D}$ & \multicolumn{4}{c}{\textbf{Network}}\\
        & \small $(\times 10^6)$ & \small $(\times 10^6)$ & \small $(\times 10^{-3})$ & \small  ResNet-50 & \small ResNet-101 & Vit-S/16 & ViT-B/16\\
                \midrule
        ImageNet & 98 & 0.128 & 1.31 & 58.1 & 58.9 & 56.3 & 57.9\\ 
		ImageNet & 98 & 0.640 & 6.54 & 65.4 & 67.2 & 64.7 & 67.2\\ 
		ImageNet & 98 & 1.281 & 13.0 & \textbf{65.9} & \textbf{67.7} &\textbf{65.4} & \textbf{68.8}\\  
        \midrule
        ImageNet & 294 & 0.128 & 0.43 & 57.5 & 59.0 & 52.9 & 56.9 \\ 
		ImageNet & 294 & 0.640 & 2.17 & 69.8 & 71.4 & 68.9 & 71.9\\ 
		ImageNet & 294 & 1.281 & 4.35 & \textbf{71.4} & \textbf{73.3} &\textbf{71.4} & \textbf{74.9}\\ 
        \bottomrule
	\end{tabular}
	}
\ifcvpr
	\caption{\textbf{Pretraining Diversity With Different Architectures Sizes}: We investigate how pretraining diversity, total computation budget and model architecture size interact for MoCoV3 when pretraining and evaluating on ImageNet. Regardless of $\mathcal{C}$ and the architecture choice, increasing pretraining diversity remains a reliable method to improve downstream results. Further, increasing model size also seems to consistently lead to better learned representations. Again, comparing pretraining diversity values only holds when the model architecture and $\mathcal{C}$ are fixed.}
    \label{tab:larger_architectures}
\fi

\end{table*}

\section{Additional Insights on Label Quantity}\label{supp:labels}
In Section~\mainref{sec:ablations} we considered how pretraining diversity affects the number of labels necessary for the best downstream ImageNet accuracies when pretrained on ImageNet (\(\mathbb{D}_{\text{SSL}} = \mathbb{D}_{\text{task}}\)) or on YFCC100M (\(\mathbb{D}_{\text{SSL}} \neq \mathbb{D}_{\text{task}}\)).
Here explore the setting where upstream and downstream data are the same, \ie, \(\mathbb{D}_{\text{SSL}} = \mathbb{D}_{\text{task}}\), and we repeat the experiment with models pretrained on various diversities on ImageNet.
Table~\ref{tab:label_quantities_supp} shows the in-domain results on ImageNet for SimCLR and MoCoV3 pretrained with ResNet-50 and ViT-B/16 backbones.
It is clear that for in-domain evaluation, the models pretrained with largest pretraining diversity always perform the best, regardless of the label quantity used for linear evaluation.
More interestingly, it is possible to achieve better performance with less labels if a model is pretrained with higher $\mathcal{D}$.
For example, for every combination of backbone and SSL method, the models pretrained with maximum diversity $\mathcal{D}=13.0\times 10^{-3}$ using 50\% of the labels outperform the models pretrained with $\mathcal{D}=2.61\times 10 ^{-3}$ with 100\% of the labels.
Thus, if models are evaluated or deployed in few-shot downstream tasks, it may be desirable to use models pretrained with the highest pretraining diversity available.

\begin{table*}[t]
\ifeccv
 \caption{\textbf{Evaluating Network Accuracy With Varied Label Quantity}: We evaluate the accuracy of networks pretrained on ImageNet with various pretraining diversities and evaluate with different labeling percentages of $\mathbb{D}_\text{task}=\text{ImageNet}$. For in-distribution data, one can get away with fewer labels using more diverse pretraining data.}
 \label{tab:label_quantities_supp}
\fi
	\centering
	\begin{subtable}{0.48\linewidth}
		\setlength\tabcolsep{0.25em} %
	\resizebox{\linewidth}{!}{%
		\begin{tabular}{cccc|cccc}
        \multicolumn{8}{c}{{\textbf{SimCLR}}}\\
		\toprule
		\multirow{2}{*}{\textbf{Network}} &  \multirow{2}{*}{$\mathbb{D}_\text{SSL}$} & $N$ & $\mathcal{D}$  & \multicolumn{4}{c}{\textbf{Accuracy$\uparrow$}}\\
		& & \small $(\times 10^6)$ & \small $(\times 10^{-3})$ & 5\% & 10\% & 50\% & 100\% \\
		\midrule
		\multirow{4}{*}{ResNet-50}  
            & \small ImageNet & 0.128 & 1.31 & 42.1 & 45.4 & 53.2 & 56.9 \\
            & \small ImageNet & 0.256 & 2.61 & 46.8 & 50.3 & 57.8 & 61.1 \\
            & \small ImageNet & 0.640 & 6.54 & 49.1 & 53.0 & 60.6 & 63.7 \\
            & \small ImageNet & 1.281 & 13.0 & \textbf{50.3} & \textbf{54.2} & \textbf{61.7} & \textbf{64.5} \\
		\midrule
		\multirow{4}{*}{ViT-B/16} 
          & \small ImageNet & 0.128 & 1.31 & 40.5 & 44.8 & 53.0 & 54.2 \\
          & \small ImageNet & 0.256 & 2.61 & 48.0 & 52.2 & 59.6 & 61.3 \\
          & \small ImageNet & 0.640 & 6.54 & 53.4 & 57.6 & 64.3 & 65.5 \\
          & \small ImageNet & 1.281 & 13.0 & \textbf{55.2} & \textbf{59.4} & \textbf{65.6} & \textbf{66.7} \\
		\bottomrule
        \\[-1.5\medskipamount]
	\end{tabular}
	}
\end{subtable}	
\hfill
\begin{subtable}{0.48\linewidth}
\setlength\tabcolsep{0.25em} %
\resizebox{\linewidth}{!}{%
	\begin{tabular}{cccc|cccc}
        \multicolumn{8}{c}{{\textbf{MoCoV3}}}\\
		\toprule
		\multirow{2}{*}{\textbf{Network}} &  \multirow{2}{*}{$\mathbb{D}_\text{SSL}$} & $N$ & $\mathcal{D}$  & \multicolumn{4}{c}{\textbf{Accuracy$\uparrow$}}\\
		& & \small $(\times 10^6)$ & \small $(\times 10^{-3})$ & 5\% & 10\% & 50\% & 100\% \\
		\midrule
		\multirow{4}{*}{ResNet-50}  
            & \small ImageNet & 0.128 & 1.31 & 43.9 & 47.4 & 55.2 & 58.1 \\
            & \small ImageNet & 0.256 & 2.61 & 49.0 & 52.7 & 60.4 & 62.9 \\
            & \small ImageNet & 0.640 & 6.54 & 51.3 & 55.4 & 63.1 & 65.4 \\
            & \small ImageNet & 1.281 & 13.0 & \textbf{52.1} & \textbf{56.2} & \textbf{63.8} & \textbf{65.9} \\
		\midrule
		\multirow{4}{*}{ViT-B/16} 
          & \small ImageNet & 0.128 & 1.31 & 47.2 & 51.3 & 58.0 & 57.9 \\
          & \small ImageNet & 0.256 & 2.61 & 53.5 & 57.6 & 63.1 & 63.7 \\
          & \small ImageNet & 0.640 & 6.54 & 57.9 & 61.7 & 66.7 & 67.2 \\
          & \small ImageNet & 1.281 & 13.0 & \textbf{59.3} & \textbf{63.0} & \textbf{68.3} & \textbf{68.8} \\
		\bottomrule
        \\[-1.5\medskipamount]
	\end{tabular}
}
\end{subtable}
\ifcvpr
 \caption{\textbf{Evaluating Network Accuracy With Varied Label Quantity}: We evaluate the accuracy of networks pretrained on ImageNet with various pretraining diversities and evaluate with different labeling percentages of $\mathbb{D}_\text{task}=\text{ImageNet}$. For in-distribution data, one can get away with fewer labels using more diverse pretraining data.}
 \label{tab:label_quantities_supp}
\fi

\end{table*}

\section{Additional details}\label{supp:details}

\subsection{SSL Methods}
\label{sec:ssl_methods}

In Section \mainref{sec:prelims} we described a general framework for self-supervised pretraining that is common to many state-of-the-art SSL methods.
Although all the methods we use in our experiments mostly follow this procedure, they do differ in loss functions as well as in certain architectural choices.
For each of the methods we use for our experiments, we describe in depth the key aspects that specifically define the SSL method and make them different from the introduced framework in Section~\mainref{sec:prelims}.
Further details on the methods can be found in their respective papers and repositories.

SimCLR~\cite{simclr}, Barlow Twins~\cite{barlow} and VICReg~\cite{vicreg} closely follow the general framework, and mainly differ in the loss function $\mathcal{L}_{\text{SSL}}$ used during optimization.
SimCLR makes use of the InfoNCE loss~\cite{oord2018representation}, which is applied to the representations of each positive pair of samples in the batch, and also incorporates negatives from the current batch.
Barlow Twins uses a loss function that makes the cross-correlation matrix between the representations of the distorted samples as close to the identity matrix as possible.
As a result, representations of correlated views of the same sample are forced to become similar, while redundancy between the components of these vectors is minimized.
The loss function used for VICReg is a combination of the mean-squared euclidean distance between representations with an additional variance and covariance term for feature decorrelation and avoiding representation collapse.

BYOL~\cite{byol}, DINO~\cite{dino} and MoCoV3~\cite{mocov3} have slightly more evident differences from the proposed framework, as they do not use shared parameters $\theta_f$ and $\theta_g$ for the feature extractor and projection head between the two augmented views. 
Instead the two augmented views pass through two different networks: a student network with feature extractor $f_{\theta_f}$ and prediction head $g_{\theta_g}$, parameterised with $\theta_f$ and $\theta_g$, and a teacher network with its own respective components $f'_{\theta_{f'}}$ and $g'_{\theta_{g'}}$ with unique parameters $\theta_{f'}$ and $\theta_{g'}$.
The weights $\theta_{f'}$ and $\theta_{g'}$ in the teacher network are an exponential moving average of the weights $\theta_f$ and $\theta_g$ in the student network.
The three methods differ in how they compute $\mathcal{L}_{\text{SSL}}$ from the representations $\mathbf{z}_A$ from the student and $\mathbf{z}_B$ from the teacher.
For BYOL, after the correlated views are passed through the two networks, an additional predictor network $q_{\theta_q}$, parameterised with $\theta_q$, tries to predict the representation of the teacher network $\mathbf{z}_B$ from the output of the student network as $q_{\theta_q}(\mathbf{z}_A)$, and the mean squared error between the teacher representation and the student prediction is minimised.
DINO performs knowledge distillation in the student by minimising the cross-entropy loss between the direct outputs $\mathbf{z}_A$ and $\mathbf{z}_B$.
MoCoV3 uses the student and teacher network to generate representations from the augmented views called \textit{queries} $\mathbf{z}_B$ and \textit{keys} $\mathbf{z}_B$, and stores the keys in a queue.
The contrastive InfoNCE loss is again used as SSL objective, and uses matching queries and keys as positive samples and the recent keys from the dictionary as negative samples.
For all three methods, a stop-gradient operator is applied after the teacher network, to avoid any weight updates in the teacher network during backpropagation.

SwAV~\cite{swav} does share weights for $f_{\theta_f}$ and $g_{\theta_g}$ between correlated views, but instead relies on additional components.
First, the representations of different views $\mathbf{z}_A$ and $\mathbf{z}_B$ are assigned to prototype vectors, resulting in \textit{codes} $\mathbf{q}_A$ and $\mathbf{q}_B$.
The prototype vectors are trainable vectors and represent the notion of clusters.
A swapped prediction problem is solved afterwards, where the code of one augmented view is predicted using the other view.
The swapped prediction problem is used to define the loss as $\mathcal{L}_{\text{SSL}}(\mathbf{z}_A, \mathbf{z}_B) = \ell (\mathbf{z}_A, \mathbf{q}_B) + \ell (\mathbf{z}_B, \mathbf{q}_A)$, where $\ell$ measures the fit between features and codes.

\subsection{Data Collection Strategies}
\label{sec:data_collection}

This section outlines the data collection strategies for our three approaches detailed in Section \mainref{sec:exp-part2}: Source, Web, and Synthetic. We base these strategies on the Base dataset $\mathbb{B}$ (introduced in \mainref{sec:exp-part2}), consisting of half of ImageNet100, totaling 65,000 samples.

\noindent\textbf{\textcolor{source}{Source Dataset}\ \ } We expand the dataset $\mathbb{B}$ by integrating the remaining half of ImageNet100, forming \ASourceIn. For \ASourceOut, we begin by selecting 100 random, non-overlapping classes from ImageNet. We then gather 65,000 corresponding samples from these classes and add them to $\mathbb{B}$.

\noindent\textbf{\textcolor{web}{Web Dataset}\ \ } We utilize three search engines—Flickr, Bing, and DuckDuckGo—to gather web samples while employing Safe-Search for content appropriateness. Our queries, based on class names, are carefully crafted to avoid ambiguity. For \AWebIn, we collect approximately 100,000 samples from ImageNet100 classes, selecting the top 65,000 for inclusion in $\mathbb{B}$. Similarly, for \AWebOut, we follow the same process for the 100 randomly selected classes from the Source dataset.

\noindent\textbf{\textcolor{synthetic}{Synthetic Dataset}\ \ } For synthetic sample generation, we employ Stable Diffusion V2.1 (SDV2.1). Using the prompt 'A photo of a \texttt{class\_name}', we generate images for each class in ImageNet100 for \ASyntheticIn and the 100 distinct classes from ImageNet for \ASyntheticOut. Each class contributes 650 images, totaling 65,000 samples. We utilize the DPMSolver++ scheduler with start and end $\beta$ values of 0.00085 and 0.012, respectively. The guidance scale is set at $w=7.5$, and each image is generated in 50 steps.

\subsection{Details on Distribution Distances}
\label{sec:domain_distances}

In Section \mainref{sec:ablations} of our study, we explored the relationship between pretraining datasets, specifically ImageNet and YFCC100M, and downstream datasets, which include ImageNet, Stanford Cars, Flowers102, OxfordIIITPets, Places365, and Food101. To quantitatively measure the distance between these datasets, we employed two distinct metrics: VisualDNA (VDNA) and the Fréchet Inception Distance (FID).

Our methodology for calculating distribution distances involved selecting a substantial number of samples from each dataset. We used $50,000$ samples from each dataset for computing the distribution distances, if the dataset is composed from less than $50,000$ samples, the whole dataset is used. This approach ensured a robust and comprehensive analysis of the dataset distributions. For the implementation of VisualDNA, we utilized two different architectures: MUGS ViT-B/16, as recommended by the original paper \cite{ramtoulaVisualDNARepresenting2023}, and DinoV2 ViT-B/16. The FID scores were computed using a standard approach with an Inception network, as detailed in \cite{fid}.

The results, as discussed in Section \mainref{sec:ablations}, revealed consistent findings across both VDNA and FID metrics. Our analysis showed that a greater distance between the upstream and downstream datasets correlated with a decrease in downstream classification accuracy.

\section{Implementation}
\label{sec:setups}

In all our experiments, we have utilized the \texttt{solo-learn} library \cite{costaSololearnLibrarySelfsupervised2022} as our main codebase. For the ImageNet100 and CIFAR100 experiments, we used to the parameters provided by \texttt{solo-learn}. In the case of ImageNet, while we began with the parameters provided in the original papers, we made slight modifications to optimize performance. These modifications included changes to the number of warmup epochs and an adjustment of the learning rate. For the YFCC100M dataset, we found the parameters optimized for ImageNet to be the most effective, whereas for TinyImageNet, we used the CIFAR100 parameters provided by \texttt{solo-learn}.

To create different fractions of each dataset, our first step involved the creation of an H5 file containing all image paths. This file is then shuffled and saved. When a specific percentage of the data is required for our SSL pretraining, we simply select the first $k\%$ of the image paths from this H5 file. Since we use a fixed computational budget, we scaled the number of epochs accordingly. This scaling is achieved by a factor of $1/k\times 100$. For example, if we utilized 10\% of the data for pretraining, we would increase the base number of epochs by a factor of $10$.

\end{document}